\documentclass{article}

\usepackage[final,nonatbib]{neurips_2023}

\usepackage[utf8]{inputenc} 
\usepackage[T1]{fontenc}    
\usepackage{hyperref}       
\usepackage{url}            
\usepackage{booktabs}       
\usepackage{amsfonts}       
\usepackage{nicefrac}       
\usepackage{microtype}      
\usepackage{xcolor}         

\usepackage{amsmath}

\DeclareMathOperator*{\argmin}{arg\,min}
\usepackage{multirow}
\usepackage{comment}
\usepackage{algorithm}
\usepackage{algorithmic,esint}
\usepackage{enumitem}
\usepackage{subfig}
\usepackage{adjustbox}

\newcommand{\model}{\textbf{DSGAS} }
\newcommand{\modelnosp}{\textbf{DSGAS}}

\newcommand{\smodelnosp}{DSGAS}

\newcommand{\ms}[2]{{#1}\scriptsize{$\pm$#2}}
\newcommand{\msone}[2]{\bf {#1}\scriptsize{$\pm$#2}}
\newcommand{\mstwo}[2]{\underline{{#1}\scriptsize{$\pm$#2}}}

\title{Unsupervised Graph Neural Architecture Search \\ with Disentangled Self-supervision}

\author{%
Zeyang Zhang\textsuperscript{1}\thanks{This work was done during the author's internship at Wechat, Tencent},\;
Xin Wang\textsuperscript{1}\thanks{Corresponding authors},\;
Ziwei Zhang\textsuperscript{1},\;
\textbf{Guangyao Shen}\textsuperscript{2},\;
\textbf{Shiqi Shen}\textsuperscript{2},\;
\textbf{Wenwu Zhu}\textsuperscript{1}\footnotemark[2]\\
\textsuperscript{1}Department of Computer Science and Technology, BNRist, Tsinghua University,\;
\textsuperscript{2}Wechat, Tencent\\
\small \texttt{zy-zhang20@mails.tsinghua.edu.cn},\;
\small \texttt{\{xin\_wang, zwzhang\}@tsinghua.edu.cn},\\
\small \texttt{\{lucasgyshen, shiqishen\}@tencent.com},\;
\small \texttt{wwzhu@tsinghua.edu.cn}\\
}

\begin{document}

\maketitle

\begin{abstract}
The existing graph neural architecture search (GNAS) methods heavily rely on supervised labels during the search process, failing to handle ubiquitous scenarios where supervisions are not available. 
In this paper, we study the problem of unsupervised graph neural architecture search, which remains unexplored in the literature.
The key problem is to discover the latent graph factors that drive the formation of graph data as well as the underlying relations between the factors and the optimal neural architectures.
Handling this problem is challenging given that the latent graph factors together with architectures are highly entangled due to the nature of the graph and the complexity of the neural architecture search process.
To address the challenge, we propose a novel \underline Disentangled \underline Self-supervised \underline Graph Neural \underline Architecture \underline Search (\modelnosp) model, which is able to discover the optimal architectures capturing various latent graph factors in a self-supervised fashion based on unlabeled graph data.
Specifically, we first design a disentangled graph super-network capable of incorporating multiple architectures with factor-wise disentanglement, which are optimized simultaneously. Then, we estimate the performance of architectures under different factors by our proposed self-supervised training with joint architecture-graph disentanglement. Finally, we propose a contrastive search with architecture augmentations to discover architectures with factor-specific expertise.
Extensive experiments on 11 real-world datasets demonstrate that the proposed \model model is able to achieve state-of-the-art performance against several baseline methods in an unsupervised manner.
\end{abstract}

\section{Introduction}

Graph neural architecture search (GNAS), aiming to automatically discover the optimal architecture for graph neural network (GNN) based on graph-structured data and task, has shown remarkable progress in enhancing the predictive power and saving human endeavors for various graph applications~\cite{zhang2021automated}. The existing GNAS methods generally follow a supervised paradigm such that they optimize the weights within architectures given a training dataset with a supervised loss (e.g., the cross entropy loss of label predictions) and estimate the architecture performance based on the validation dataset with supervision signals. For example, the label prediction accuracy is adopted for architecture ranking during the neural architecture search process~\cite{li2020autograph, gao2020graph,wei2021pooling}. As a result, supervised labels become indispensable for applying the existing GNAS methods.  

However, ground-truth labels in reality may be extremely scarce or hardly available in many graph applications. 
For example, a variety of biological problems require a significant amount of human labors and time costs in clinical tests to obtain labels for supervision~\cite{martin2017much,paul2021artificial,sun202290}.
As the existing GNAS approaches heavily rely on supervised labels for weight training and architecture evaluation, 
they will suffer from performance deterioration in unsupervised settings, 
failing to discover optimal architectures in the scenarios where labels are scarce or not available.

In this paper, we study unsupervised graph neural architecture search, i.e., discovering optimal GNN architectures without labels for graph-structured data, which remains unexplored in the literature. 
The key problem lies in two important aspects: i) discover the latent graph factors that drive the formation process of graph data~\cite{fan2019graph,ma2019disentangled,yang2020factorizable,li2022disentangled, cooray2022graph,xiao2022decoupled,li2021disentangled}; 
ii) capture the underlying relations between the factors and the optimal neural architectures.
For instance, a molecular graph may consist of groups of atoms as well as bonds representing different functional units~\cite{ying2018hierarchical}, requiring different optimal neural architectures to make accurate predictions. 

Nevertheless, solving the problem is highly non-trivial and challenging 
given that the hidden factors are entangled in the graph and very difficult to capture, 
e.g., a social network may contain several communities originating from various interests (e.g., sports, games, etc.)~\cite{ying2019gnnexplainer,yang2020factorizable}, with the nodes and edges belonging to different communities mixing together.  
Moreover, the architectures with different functional factors are also entangled within the weight-sharing super-network~\cite{chu2020fair,chu2021fairnas,cha2022supernet}, resulting in inaccurate architecture performance estimations under different hidden factors. 

To tackle the challenge, we propose a novel unsupervised graph neural architecture search method, i.e., Disentangled Self-supervised Graph Neural Architecture Search (\modelnosp)\footnote{The codes are available at \href{https://github.com/wondergo2017/dsgas}{Github}.}. Given graph data without supervised labels, our proposed \model model can discover the optimal architectures capturing multiple latent factors in a self-supervised fashion. 
In particular, we first design a disentangled graph super-network, where multiple architectures are disentangled for simultaneous optimization w.r.t various latent factors.
Then, we propose a self-supervised training with joint architecture-graph disentanglement, which disentangles architectures and graphs within a common latent space. The super-network is trained through a routing mechanism between architectures, graphs and self-supervised tasks, to obtain an accurate estimation of the architecture performance under each latent factor. 
Finally, we propose a contrastive search with architecture augmentations, where a novel architecture-level instance discrimination task is introduced to discover architectures with distinct capabilities of capturing various factors in a self-supervised fashion. Extensive experiments show that the proposed \model model is able to significantly outperform the state-of-the-art GNAS baselines under both unsupervised and semi-supervised settings. Detailed ablation studies and analyses also demonstrate that \model is able to discover effective architectures with our proposed disentangled self-supervision designs.
The contributions of this paper are summarized as follows:
\begin{itemize}[leftmargin=0.5cm]
    \item We are the first to study the problem of unsupervised graph neural architecture search and propose the Disentangled Self-supervised Graph Neural Architecture Search (\modelnosp) model capable of discovering the optimal architectures without labels, to the best of our knowledge. 
    \item We introduce three novel modules, i) disentangled graph architecture super-network, ii) self-supervised training with joint architecture-graph disentanglement and iii) contrastive search with architecture augmentations, which can discover the optimal architectures capturing various graph latent factors with disentangled self-supervision.
    \item Extensive experiments on 11 real-world graph datasets show that our proposed method \model is able to discover effective graph neural architectures without supervised labels and significantly outperform the state-of-the-art baselines in both unsupervised and semi-supervised settings.
\end{itemize}

\section{Preliminaries and Problem Formulation}
\paragraph{Graph Neural Architecture Search}
Denote the graph space as $\mathcal{G}$ and the label space as $\mathcal{Y}$. A graph neural network can be denoted as a function $f_{\alpha,w}:\mathcal{G} \rightarrow \mathcal{Y}$, which is characterized by architecture parameters $\alpha \in \mathcal{A}$ and learnable weights $w \in \mathcal{W}$ given an architecture space $\mathcal{A}$ and a weight space $\mathcal{W}$. Graph neural architecture search (GNAS) aims at automating the design of graph neural architectures, i.e., obtaining the best-performed architectures by searching $\alpha$. As $\alpha$ is usually instantiated as selecting GNN operations, e.g., GCN~\cite{kipf2016semi}, GAT~\cite{velivckovic2018graph}, GIN~\cite{xu2018powerful}, we also call $\alpha$ as operation choices for brevity. Generally, GNAS solves the bi-level optimization problem~\cite{elsken2019neural} :
\begin{align}
\alpha^{*} &=\argmin_{\alpha \in \mathcal{A}} \mathcal{L}_{\text{val}}(\alpha, w^{*}(\alpha)),\label{eq:alpha} \\ 
\text{ s.t. } w^{*}(\alpha) &=\argmin_{w \in \mathcal{W}(\alpha)} \mathcal{L}_{\text{train}}(\alpha, w), \label{eq:weight}
\end{align}
where $\mathcal{L}_{\text{train}}$ and $\mathcal{L}_{\text{val}}$ denotes the loss of the predictions of the architecture $f_{\alpha,w}(\cdot)$ against supervised labels on training and validation datasets. The optimization problem can be viewed as having two objectives that Eq.\eqref{eq:weight} aims to obtain accurate architecture performance estimation, and Eq.\eqref{eq:alpha}  aims to search the best-performed architectures. To avoid the cost of training from scratch for each architecture, the super-network~\cite{pham2018efficient,liu2018darts} arises as a commonly adopted technique to obtain faster architecture performance estimation, where the architecture candidates are viewed as sub-networks of the super-network, and their weights are shared during the training process.  

\paragraph{Unsupervised Graph Neural Architecture Search}
We consider the problem of unsupervised graph neural architecture where labels, which are adopted for the performance estimation and the search process in the supervised GNAS, are not accessible. 
The problem of unsupervised GNAS can be formulated as optimizing an architecture generator that is able to discover powerful architectures by exploiting inherent graph properties without labels, i.e., $\mathcal{G} \mapsto f_{\alpha,w}$ instead of $(\mathcal{G},\mathcal{Y}) \mapsto f_{\alpha,w}$ as done by supervised GNAS methods. Then, the discovered architectures $f_{\alpha,w}(\cdot)$ can be utilized in downstream tasks, e.g., finetuning the weights $w$ or the operation choices $\alpha$, or extra shallow classifiers for further prediction. 

\begin{figure}
    \centering
    \includegraphics[width=1\textwidth]{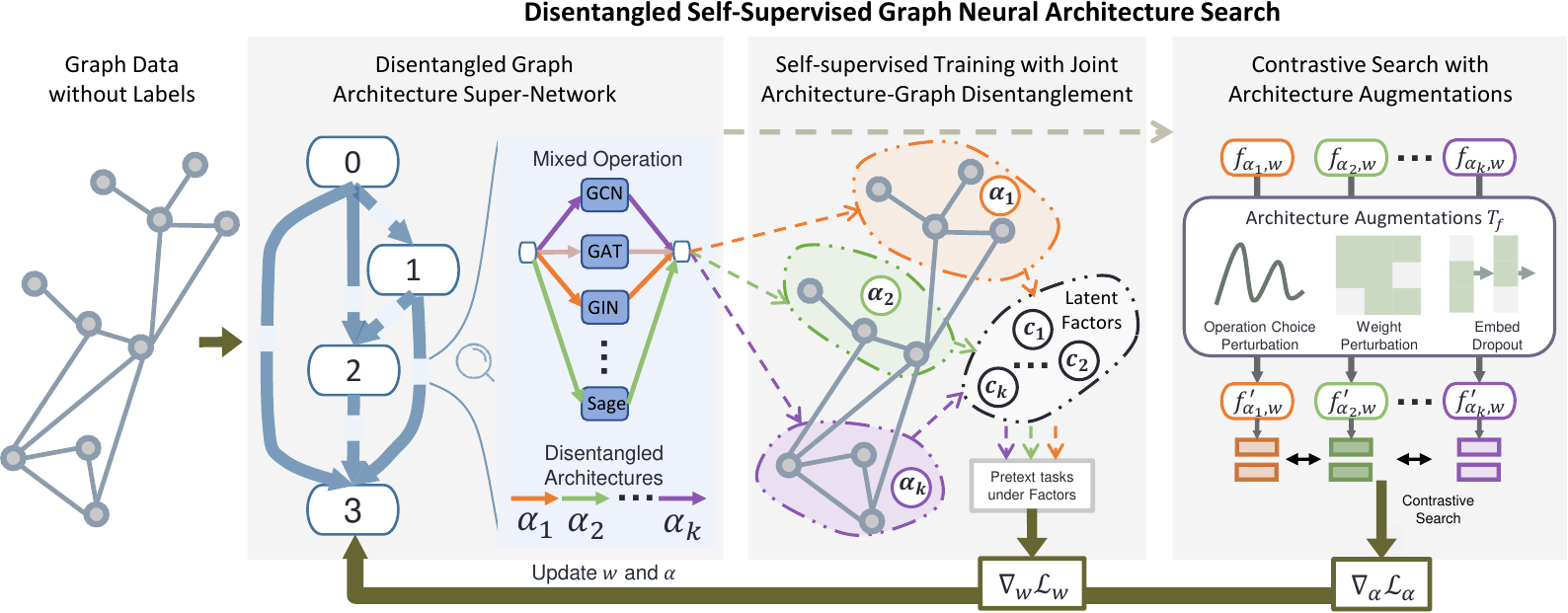}
    \caption{The framework of Disentangled Self-supervised Graph Neural Architecture Search (\modelnosp), including the following three key components: 1) Disentangled graph architecture super-network enables multiple architectures to be disentangled and optimized simultaneously in an end-to-end manner. 2) Self-supervised training with joint architecture-graph disentanglement estimates the performance of architectures under various latent factors by considering the relationship between architectures, graphs and factors. 3) Contrastive search with architecture augmentations encourages and discovers architectures with distinct capabilities of capturing factors. (Best viewed in color) }
\end{figure}

\section{Disentangled Self-supervised Graph Neural Architecture Search}

In this section, we introduce Disentangled Self-supervised Graph Neural Architecture Search (\modelnosp) to search architectures without labels, by proposing three key components: disentangled graph architecture super-network, self-supervised training with joint architecture-graph disentanglement, and contrastive search with architecture augmentations.

\subsection{Disentangled Graph Architecture Super-Network}
\label{sec:supernet}
To discover architectures that potentially have optimal performance, we resort to guiding the search towards architectures' capabilities of capturing the inherent graph factors, which are shown important in the graph formation~\cite{fan2019graph,ma2019disentangled}.
As architectures may expert in different graph factors, we propose a disentangled graph architecture super-network to incorporate $K$ different architectures to be estimated and searched w.r.t factors simultaneously, where the hyperparameter $K$ denotes the number of factors.

\paragraph{Disentangled Super-Network Layer} For each super-network layer, we adopt $K$ mixed operations parameterized by different $\mathbf{\alpha}$ to learn $K$-chunk graph representations:
\begin{equation}
\begin{aligned}
\mathbf{H}_k \leftarrow \overline{\text{GNN}}_{\mathbf{\alpha}_k}\left(\mathbf{H}, \mathbf{A}\right),
\end{aligned}
\end{equation}
where $\mathbf{A}$ is the adjacency matrix of the graph, 
$\mathbf{H}$ 
denotes the input graph representations, 
and $\overline{\text{GNN}}_{\alpha_k}(\cdot)$ denotes the mixed GNN operations parameterized by $\mathbf{\alpha}_k$. For the convenience of differentiable optimization, we adopt continuous parameterization and weight-sharing mechanism~\cite{liu2018darts} to implement the mixed operations:
\begin{equation}\label{eq:mix_op}
\overline{\text{GNN}}_{\mathbf{\alpha}_k}(\mathbf{H},\mathbf{A})=\sum_{i=1}^{|\mathcal{O}|} \alpha_{k,i}\text{GNN}_i(\mathbf{H},\mathbf{A}),
\end{equation}
where $|\mathcal{O}|$ is the number of GNN operation choices, $\alpha_{k,i}=\frac{\exp(\theta_{\alpha_{k,i}})}{\sum_j \exp(\theta_{\alpha_{k,j}})}$ denotes the probability of the $i$-th operation for the $k$-th architecture $\alpha_k$, and $\theta$ is learnable parameters.

\paragraph{Overall Disentangled Super-Network} The overall super-network is constructed in the form of a directed acyclic graph (DAG) with an ordered sequence of disentangled super-network layers. More details about the DAG construction and the according GNN operations for each disentangled super-network layer are included in Appendix. The output of the last layer $\mathbf{Z}=[\mathbf{H}_1,\mathbf{H}_2,\dots,\mathbf{H}_K]$ describes the various aspects of the graphs and serves as the final graph representations, which can be utilized or finetuned in downstream tasks. In this way, the architectures' operation choices $\mathbf{\alpha}=[\mathbf{\alpha}_1,\mathbf{\alpha}_2,\dots,\mathbf{\alpha}_K]$ and weights $w$ are incorporated in one super-network. For brevity, we use $f_{\alpha_k,w}(\cdot)$ to denote the $k$-th architecture induced from the super-network. Note that the design of $K$ operation choices alleviates the entanglement of architectures by providing more flexible choices of paths~\cite{guo2020single} in the super-network. For instance, the `mean' operation captures structural properties while the `max' operation captures representative elements~\cite{xu2018powerful}, and in this case, our design can capture both of them by choosing corresponding operations to learn respective representations instead of choosing only one of them which may conflict each other.

\subsection{Self-supervised Training with Joint Architecture-Graph Disentanglement}
\label{sec:ssl_train}
Inspired by graph self-supervised learning, we utilize graph pretext tasks to measure the architectures' capabilities of capturing latent factors. Predictive pretext tasks~\cite{wu2021self}, for example, design a pseudo label generator $s(\cdot)$, and optimize the prediction probability\footnote{We take graph classification as an example for simplicity, while the case of node classification can be easily extended.} $p(s(\mathcal{G}_i)|\mathcal{G}_i)$. However, the tasks usually take holistic views of the graphs and neglect the entanglement of the latent factors, which may lead to suboptimal performance estimation. Therefore, to disentangle the factors, we transform the probability into the expectation of multiple subtasks under different latent factors by the Bayesian formula:
\begin{equation}\label{eq:ssl-train}
p(s(\mathcal{G}_i)|\mathcal{G}_i) = \mathbb{E}_{p(k|\mathcal{G}_i)} p(s(\mathcal{G}_i)|\mathcal{G}_i,k),
\end{equation}
where $p(k|\mathcal{G}_i)$ denotes the probability of latent factor $k$ given the $i$-th graph instance $\mathcal{G}_i$, and $p(s(\mathcal{G}_i)|\mathcal{G}_i,k)$ denotes the pretext task under $k$-th latent factor. An intuitive explanation of Eq.~\eqref{eq:ssl-train} is that it first infers the latent factors and then conducts factor-specific self-supervised training to capture the latent factors, which we describe in detail as follows.   

\paragraph{Architecture-aware Latent Factor Inference} Directly modeling $p(k|\mathcal{G}_i)$ is difficult as we do not know prior what GNN encoders are suitable for inferring the latent factors. Intuitively, one solution is to get the architectures being searched involved in the inference stage. By the Bayesian formula, we factorize the probability w.r.t architecture choices:
\begin{equation}
p(k|\mathcal{G}_i)=\mathbb{E}_{p(\alpha_j | \mathcal{G}_i)} p(k|\mathcal{G}_i,\alpha_j),
\end{equation}
where $p(\alpha_j|\mathcal{G}_i)$ is a prior distribution, and we adopt a uniform distribution for simplicity. Then we can model the probability distributions of latent factors given the graph $\mathcal{G}_i$ by utilizing the architectures being searched:
\begin{equation}
p(k|\mathcal{G}_i,\alpha_j)=\frac{\exp \phi(\mathbf{z}_{i,j}||\text{Enc}(\alpha_j), \mathbf{c}_k)}{\sum_{m=1}^{K} \exp \phi(\mathbf{z}_{i,j}||\text{Enc}(\alpha_j), \mathbf{c}_m)},
\end{equation}
where $\mathbf{c}_k$ is a learnable vector to represent the $k$-th latent factor, and $\mathbf{z}_{i,k}=f_{\alpha_k,w}(\mathcal{G}_i)$ denotes the graph representations output by the $k$-th architecture for the graph $\mathcal{G}_i$. $\text{Enc}(\cdot)$ denotes architecture encoding techniques to obtain embeddings of operation choices $\alpha$ so that the structural properties and correlations of the neural architectures can be considered~\cite{yan2020does,luo2018neural}. 

\paragraph{Factor-aware Graph Self-Supervised Learning} Under the $k$-th latent factor, we leverage the corresponding architectures $f_{\alpha_k,w}(\cdot)$ for conducting the factor-specific pretext tasks as $p(s(\mathcal{G}_i)|\mathcal{G}_i,k)$ to estimate the capturing capabilities of architectures under various factors. The overall objective is to maximize Eq.\eqref{eq:ssl-train}, and the loss can be calculated by 
\begin{equation}
\frac{1}{N}\sum_i -\log\mathbb{E}_{p(k|\mathcal{G}_i)}\Bigl(p(s(\mathcal{G}_i)|\mathcal{G}_i,k)\Bigr) \leq \frac{1}{N} \sum_i \mathbb{E}_{p(k|\mathcal{G}_i)} \Bigl(-\log p(s(\mathcal{G}_i)|\mathcal{G}_i,k)\Bigr),
\end{equation}
where $N$ is the number of samples and the upper bound is obtained by Jensen's Inequality. Then we can generalize our method to other graph self-supervised tasks with specially-designed task loss functions by defining $-\log p(s(\mathcal{G}_i)|\mathcal{G}_i,k)$ as the task loss function $l(f_{\alpha_k,w},\mathcal{G}_i)$ for the graph $\mathcal{G}_i$ under the $k$-th factor, and calculate the loss by
\begin{equation}
 \mathcal{L}_{w} = \frac{1}{N} \sum _i \mathbb{E}_{p(k|\mathcal{G}_i)}\Bigl(l(f_{\alpha_k,w},\mathcal{G}_i)\Bigr).
\end{equation}
In this way, the disentangled architectures in Sec.~\ref{sec:supernet} coupled with factors disentangled from graph data can be routed pairwisely, and trained with factor-specific self-supervision to obtain more accurate performance estimation under each factor. Similar to ~\cite{liu2018darts}, the super-network weights are updated with $w = w - \lambda_w \nabla_w \mathcal{L}_{w}$ to obtain the weights that can represent the architectures' capabilities. 

\subsection{Contrastive Search with Architecture Augmentations}
\label{sec:ssl_search}
In this section, we focus on encouraging the disentanglement of architectures and searching architectures with distinct capabilities of capturing different factors. The main insight of our proposed search method is intuitively based on the following two observations shown in the literature: 1) As architectures similar in operation choices and topologies have similar capabilities of capturing semantics for downstream tasks~\cite{yan2020does,ying2019bench,zela2022surrogate}, slight modifying the architecture will have a slight influence on its capability. 2) Since different GNN architectures expert in different downstream tasks~\cite{you2020design}, the architectures searched for different disentangled latent factors are expected to have dissimilar capabilities under different factors. 

\paragraph{Contrastive Search} Inspired by self-supervised contrastive learning~\cite{liu2021self,wang2022chaos} that capture discriminative features by pulling similar instances together and pushing dissimilar instances away in the latent space, we propose an architecture-level instance discrimination task to encourage the architectures to capture various latent factors. The task is defined as   
\begin{equation}
\begin{aligned}
p(s(\alpha_k) | \mathcal{G}_i,\alpha_k)=\frac{\exp \phi(\mathbf{z}_{i,k}, \mathbf{z}_{i,s(\alpha_k)}^{\prime})}{\sum_{j=1}^N \exp \phi(\mathbf{z}_{i,j}, \mathbf{z}_{i,s(\alpha_j)}^{\prime})},
\end{aligned}
\end{equation}
\begin{equation}
\mathbf{z}_{i,k}=f_{\alpha_k,w}(\mathcal{G}_i)
,\mathbf{z}_{i,k}^{\prime}=T_f(f_{\alpha_k,w})(\mathcal{G}_i),
\end{equation}
where $s(\alpha_k)$ is assigned to $k$ as surrogate labels for the architecture, $\phi(\cdot)$ calculates the similarity of two embeddings and $T_f(\cdot)$ denotes architecture augmentations that transform an architecture to another architecture with similar capabilities capturing factors, i.e., $f_{\alpha_k,w} \mapsto f^{\prime}_{\alpha_k,w}$. Then the loss function can be calculated by
\begin{equation}
\mathcal{L}_{\alpha}=\sum_i -\log \mathbb{E}_{p(\alpha_k|\mathcal{G}_i)}p(s(\alpha_k)| \mathcal{G}_i,\alpha_k).
\end{equation}
Similar to ~\cite{liu2018darts}, the architecture parameters are updated with $\alpha = \alpha - \lambda_\alpha \nabla_{\alpha} \mathcal{L}_{\alpha}$ to search architectures with better capabilities of capturing factors. 

\paragraph{Architecture Augmentations} To create various views of architectures, we design three basic architecture augmentations from the perspectives of architecture operation choices $\mathbf{\alpha}$, architecture weights $w$ and the internal embeddings $\mathbf{H}$: 
\begin{itemize}[leftmargin=0.5cm]
    \item Operation Choice Perturbation. This augmentation randomly reshapes the distributions of the mixed operations by altering the temperature in the softmax function in Eq.~\eqref{eq:mix_op} with
    \begin{equation}
    \alpha_{k,i}=\frac{\exp(\theta_{\alpha_{k,i}}/\tau)}{\sum_j \exp(\theta_{\alpha_{k,j}}/\tau)},
    \end{equation}
    where the temperature $\tau$ is sampled from a uniform distribution $\mathcal{U}([1/r_1,r_1])$, and $r_1 \geq 1$ is a hyper-parameter controlling the perturbation degree.  
    \item Weight Perturbation. This augmentation randomly adds Gaussian noises $\epsilon \sim \mathcal{N}(0,\sigma^2)$ to $r_2$\% of the architecture weights $w$, where $r_2$ controls the perturbation ratio, and $\sigma^2$ is the standard deviation of the weights. 
    \item Embedding Dropout. This augmentation randomly drops $r_3$\% of the embeddings $\mathbf{H}$ output from the mixed operations, where $r_3$ controls the dropout ratio.
\end{itemize}

Note that these architecture augmentations can further be randomly composed to generate mixed augmentations. 

\section{Experiments}
In this section, we conduct experiments on 8 real-world datasets with unsupervised settings to verify the design of our method. We also include detailed ablation studies to analyze the effectiveness of each component, and 3 real-world datasets in semi-supervised settings to show that our method can alleviate the label scarcity issues by pretraining the super-network. 

\paragraph{Baselines} We compare our method with 11 baselines from the following two different categories.
\begin{itemize}[leftmargin=0.5cm]
    \item \textbf{Manually designed GNNs.} We include five representative GNNs as our baselines, i.e., GCN~\cite{kipf2016semi}, GAT~\cite{velivckovic2018graph}, GIN~\cite{xu2018powerful}, GraphSage~\cite{hamilton2017inductive}, GraphConv~\cite{morris2019weisfeiler} and a simple baseline MLP, which constitutes our search space. For graph-level classification tasks, we adopt global mean pooling for each layer and concatenation to obtain the graph representations for these baselines. 
    \item \textbf{Graph neural architecture search.} We include representative GNAS baselines GraphNAS~\cite{gao2020graph}, PAS~\cite{wei2021pooling} and GASSO~\cite{qin2021graph}, where PAS is specially designed for graph-level classification tasks by searching the pooling operations, and GASSO is specially designed for node-level classification tasks by searching the graph structures simultaneously. We also include two classical NAS baselines, random search and DARTS~\cite{liu2018darts}. As these baselines are not specially designed for graphs, we adopt our search space for these baselines. 
\end{itemize}

\begin{table*}[htbp]
\centering
\caption{Summary of dataset statistics. Unsup./Semi. denotes Unsupervised and Semi-supervised settings. Graph/Node denotes graph and node classification tasks. ACC/AUC denotes Accuracy and ROC-AUC evaluation metrics.}
\adjustbox{max width=0.98\textwidth}{
\begin{tabular}{cccccccccccc}
\toprule
\textbf{Datasets}           & \textbf{MUTAG} & \textbf{IMDB-B} & \textbf{PROTEINS} & \textbf{DD} & \textbf{Computers} & \textbf{Photos} & \textbf{CS} & \textbf{Physics} & \textbf{OGBG-Molhiv} & \textbf{OGBN-Arxiv} & \textbf{Wechat-Video} \\ \midrule
\textbf{Setting}            & Unsup.              & Unsup.                    & Unsup.                 & Unsup.           & Unsup.                        & Unsup.                      & Unsup.                    & Unsup.                         & Semi.                    & Semi.                   & Semi.                     \\
\textbf{Task}               & Graph              & Graph                    & Graph                 & Graph           & Node                        & Node                      & Node                    & Node                         & Graph                    & Node                   & Node                     \\
\textbf{Evaluation Metrirc} & ACC              & ACC                    & ACC                 & ACC           & ACC                        & ACC                      & ACC                    & ACC                         & AUC                    & ACC                   & AUC                     \\
\textbf{\# Graphs}          & 188            & 1,000                 & 1,113              & 1,178        & 1                        & 1                      & 1                    & 1                         & 41,127                & 1                   & 1                     \\
\textbf{\# Avg. Nodes}      & 17             & 19                   & 39                & 284         & 13,752                    & 7,650                   & 18,333                & 34,493                     & 26                   & 169,343              & 60,774                 \\
\textbf{\# Avg. Edges}      & 56             & 211                  & 183               & 1,714        & 505,474                   & 245,812                 & 182,121               & 530,417                    & 28                   & 2,484,941             & 3,182,156               \\
\textbf{\# Features}        & 7              & 1                    & 3                 & 89          & 767                      & 745                    & 6,805                 & 8,415                      & 300                  & 128                 & 512                   \\
\textbf{\# Classes}         & 2              & 2                    & 2                 & 2           & 10                       & 8                      & 15                   & 5                         & 2                    & 40                  & 13                    \\ \bottomrule
\end{tabular}
\label{tab:data}
}
\end{table*}

\paragraph{Datasets} 
For unsupervised settings, we conduct experiments on four graph-level classification datasets including PROTEINS~\cite{dobson2003distinguishing}, DD~\cite{shervashidze2011weisfeiler}, MUTAG~\cite{debnath1991structure}, IMDB-B~\cite{yanardag2015deep} from TUDataset~\cite{morris2020tudataset} and four node-level classification datasets Coauthor CS, Coauthor Physics from the Microsoft Academic Graph~\cite{sinha2015overview}, Amazon Computers, Amazon Photos from the Amazon Co-purchase Graph~\cite{mcauley2015image}. For semi-supervised settings, we adopt three real-world datasets, OGBG-Molhiv, OGBN-Arxiv~\cite{hu2020open} and Wechat-Video\footnote{https://algo.weixin.qq.com/}.  
The datasets cover various graph-related fields including small molecules, bioinformatics, social networks, e-commerce networks, and academic coauthorship networks. The statistics are summarized in Table~\ref{tab:data}.

\paragraph{Super-network} 
We briefly introduce the super-network construction as follows. The super-network consists of two parts, the operation pool and the directed acyclic graph (DAG). The operation pool includes several node aggregation operations (e.g., GCN, GIN, GAT, etc.), graph pooling operations (e.g., SortPool, AttentionPool, etc.), and layer merging operations (e.g., MaxMerge, ConcatMerge, etc.). The DAG determines how the operations are connected to calculate the graph representations for the subsequent classification tasks.

More details of the experiments are provided in the Appendix, including additional experiments and analyses, experimental setups, configurations, and implementation details. 

\begin{table}[t]
\small
\centering
\caption{The results (accuracy\%) of all the methods on the real-world datasets in unsupervised settings. Numbers after the $\pm$ signs represent standard deviations. The best results are in bold and the second-best results are underlined. As the search space of GASSO and PAS do not suit the graph-level and node-level tasks respectively, we omit their results. }
\label{tab:main}
\adjustbox{max width=\textwidth}{
\begin{tabular}{ccccccccc}
\toprule
\textbf{Data}   & \multicolumn{4}{c}{\textbf{Graph Classification}}                             & \multicolumn{4}{c}{\textbf{Node Classification}}                               \\
\textbf{Method} & \textbf{PROTEINS} & \textbf{DD}       & \textbf{MUTAG}    & \textbf{IMDB-B}   & \textbf{CS}       & \textbf{Computers} & \textbf{Physics}  & \textbf{Photo}    \\ \midrule
GCN             & \ms{72.8}{0.7}    & \ms{77.0}{0.9}    & \ms{78.6}{1.6}    & \ms{63.5}{0.8}    & \ms{93.0}{0.3}    & \mstwo{86.0}{0.4}  & \ms{95.7}{0.1}    & \ms{90.8}{0.6}    \\
GAT             & \ms{72.3}{0.9}    & \mstwo{77.5}{0.7} & \ms{78.0}{0.8}    & \ms{54.4}{1.7}    & \mstwo{93.4}{0.3} & \ms{85.8}{0.3}     & \ms{95.6}{0.1}    & \ms{91.4}{0.6}    \\
GIN             & \ms{72.6}{0.4}    & \ms{77.3}{0.7}    & \ms{86.3}{1.7}    & \ms{70.7}{0.5}    & \ms{93.1}{0.3}    & \ms{76.7}{0.5}     & \ms{95.3}{0.1}    & \ms{91.1}{0.7}    \\
GraphSage       & \ms{72.9}{0.7}    & \ms{77.1}{0.4}    & \ms{78.3}{1.6}    & \ms{53.0}{2.1}    & \ms{93.2}{0.3}    & \ms{78.4}{0.4}     & \ms{95.4}{0.1}    & \ms{89.2}{0.7}    \\
GraphConv       & \ms{72.1}{0.6}    & \ms{77.3}{0.6}    & \mstwo{87.2}{1.4} & \mstwo{71.1}{0.6} & \ms{93.1}{0.3}    & \ms{74.7}{0.7}     & \ms{95.3}{0.1}    & \ms{91.5}{0.5}    \\
MLP             & \ms{70.5}{0.4}    & \ms{76.1}{0.7}    & \ms{74.8}{1.1}    & \ms{50.3}{0.6}    & \ms{91.5}{0.4}    & \ms{56.6}{0.3}     & \ms{94.6}{0.1}    & \ms{87.4}{0.8}    \\ \midrule
Random          & \ms{74.5}{0.9}    & \ms{74.8}{1.3}    & \ms{82.1}{2.8}    & \ms{69.0}{2.1}    & \ms{92.9}{0.3}    & \ms{84.8}{0.4}     & \ms{95.4}{0.1}    & \ms{91.1}{0.6}    \\
DARTS           & \ms{73.6}{0.9}    & \ms{75.7}{0.9}    & \ms{86.5}{2.3}    & \ms{70.4}{0.6}    & \ms{92.8}{0.3}    & \ms{79.7}{0.5}     & \ms{95.2}{0.1}    & \ms{91.5}{0.6}    \\
GraphNAS        & \ms{73.6}{0.7}    & \ms{75.2}{0.9}    & \ms{77.5}{0.7}    & \ms{62.7}{1.3}    & \ms{91.6}{0.3}    & \ms{69.0}{0.6}     & \ms{94.5}{0.1}    & \ms{89.3}{0.7}    \\
PAS             & \mstwo{74.6}{0.3} & \ms{76.5}{0.9}    & \ms{84.0}{1.6}    & \ms{64.6}{13.8}   & -                 & -                  & -                 & -                 \\
GASSO           & -                 & -                 & -                 & -                 & \ms{93.1}{0.3}    & \ms{84.9}{0.4}     & \mstwo{95.7}{0.1} & \mstwo{92.0}{0.3} \\ \midrule
\model          & \msone{76.0}{0.2} & \msone{78.4}{0.7} & \msone{88.7}{0.7} & \msone{72.0}{0.5} & \msone{93.5}{0.2} & \msone{86.6}{0.4}  & \msone{95.7}{0.1} & \msone{93.3}{0.3} \\ \bottomrule
\end{tabular}
}
\end{table}

\subsection{Main Results}

\paragraph{Unsupervised Settings} From the experimental results summarized in Table \ref{tab:main}, we have the following observations: 1) {\it GNNs perform differently on various datasets.} The best GNN baselines for these datasets are GraphSage, GAT, GraphConv, GraphConv, GAT, GCN, GCN, GraphConv successively, and the performance varies greatly across datasets and GNNs. It verifies that no GNN architecture is dominant for all datasets, which is consistent with the literature~\cite{you2020design} and shows the demand for automated GNN architecture designing based on the data characteristics to obtain the optimal representations. 2) {\it Most GNAS baselines fail in the unsupervised setting.} Since the existing GNAS baselines highly rely on supervised signals to search the architectures, they inherently do not suit the unsupervised settings. As an ad hoc remedy for the existing GNAS baselines in unsupervised settings, we substitute the supervised metrics with the self-supervised ones during the searching process as simple extensions. However, these GNAS baselines, contrary to supervised settings, do not guarantee better performance than manually designed GNNs for all datasets. For example, all GNAS baselines even perform worse than manually designed GNNs on DD dataset and do not have significant improvements on most datasets. The reasons behind might be that simply using graph self-supervised metrics does not consider the entanglement of architectures and factors, leading to inaccurate estimation of the architectures' capabilities. 3) {\it Our method has significant improvements over the baselines on most datasets.} Compared with manually designed GNN baselines, \model has a performance improvement of 3.1\% on PROTEINS and over 1\% on most datasets. We contribute this to its ability of automatically tailoring GNNs for various datasets, showing its effectiveness of automatic GNN designing. \model also significantly surpasses GNAS baselines, showing its superiority in graph neural architecture search in unsupervised settings, which benefits from the design of discovering architectures that can capture various graph factors in a self-supervised fashion. 

\begin{figure}
    \centering
    \includegraphics[width=0.85\textwidth]{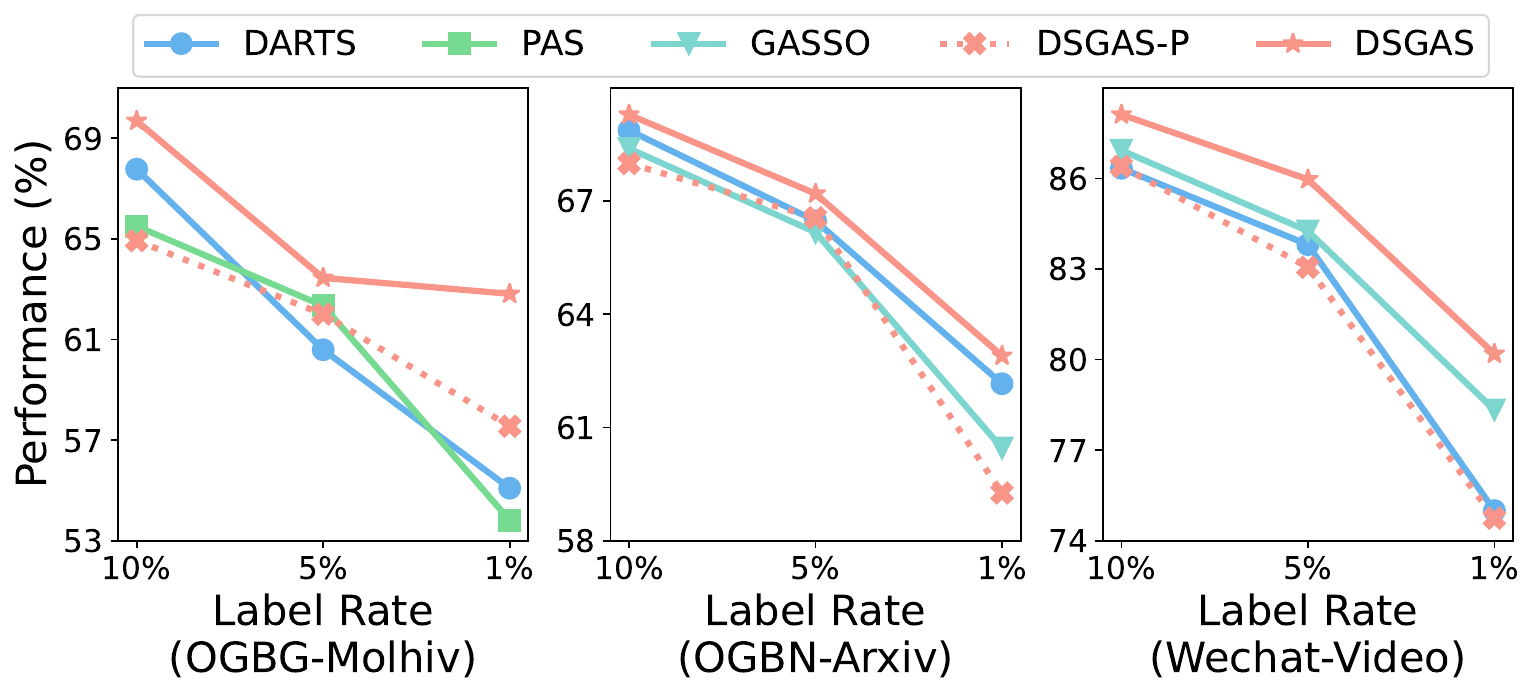}
    \caption{The performance of GNAS methods on real-world datasets under semi-supervised settings, where DSGAS-P denotes DSGAS without pretraining. The results are averaged by five random runs. (Best viewed in color)}
    \label{fig:pretrain}
\end{figure}

\paragraph{Semi-supervised Settings} From Figure~\ref{fig:pretrain}, we have the following observations: Compared with the baselines, \model significantly alleviates the performance drop when the number of available supervised labels is fewer, which verifies that our method fully exploits latent factors inside graph data and boost the supervised architecture search stage by warming up the weights and architecture parameters of the super-network. Its significant improvement over the ablated version DSGAS-P also verifies the effectiveness of pretraining the super-network by the proposed modules of self-supervised training with joint architecture-graph disentanglement and contrastive search with architecture augmentations. For example, our model with the pretraining stage has an absolute improvement of 5\% on Wechat-Video dataset with 1\% labels compared with the ablated version without the pretraining stage, which shows the effects of pretraining the super-networks on alleviating the label scarcity issues. 
In comparison, the performance of other baselines decays significantly more than our method, showing that current GNAS can not tackle scenarios with scarce labels. For example, on OGBG-Molhiv, PAS is the best baseline while the worst with 10\% and 1\% labels respectively, which may due to the inaccurate performance estimation with scarce labels. The phenomenon further strengthens the necessity of designing effective unsupervised GNAS methods. 

\subsection{Additional Experiments}

\begin{figure}
    \centering
    \subfloat[]{\includegraphics[width=0.48\textwidth]{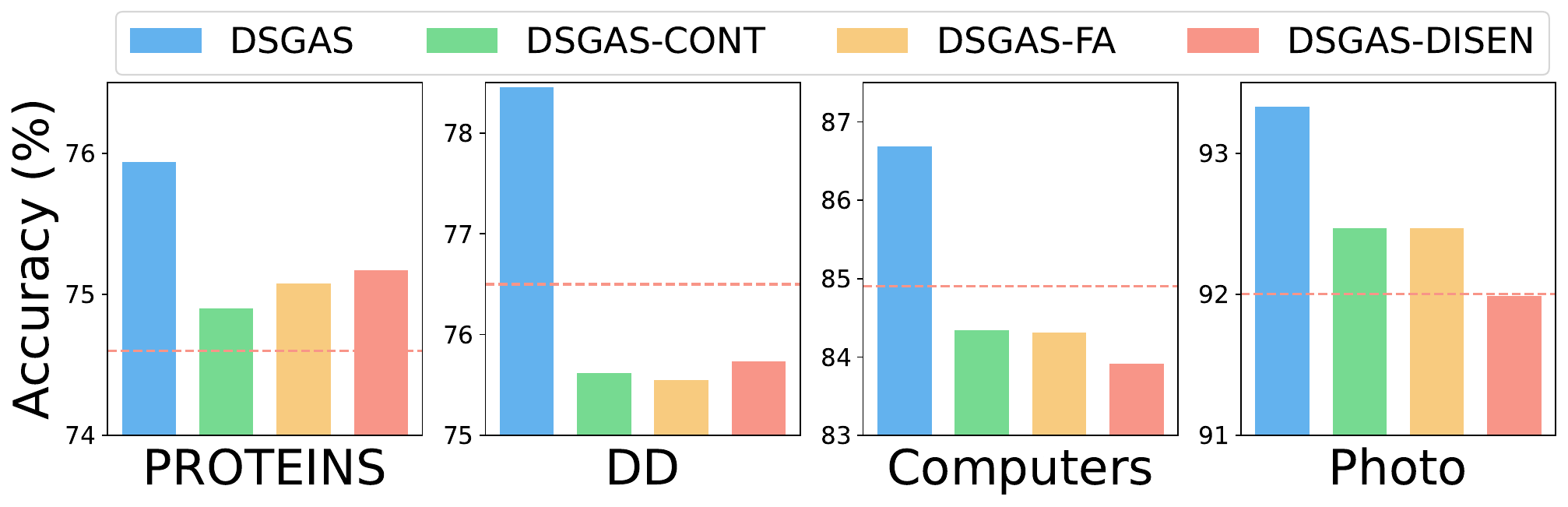}
    \label{fig:ablation}
    }
    \subfloat[]{\includegraphics[width=0.48\textwidth]{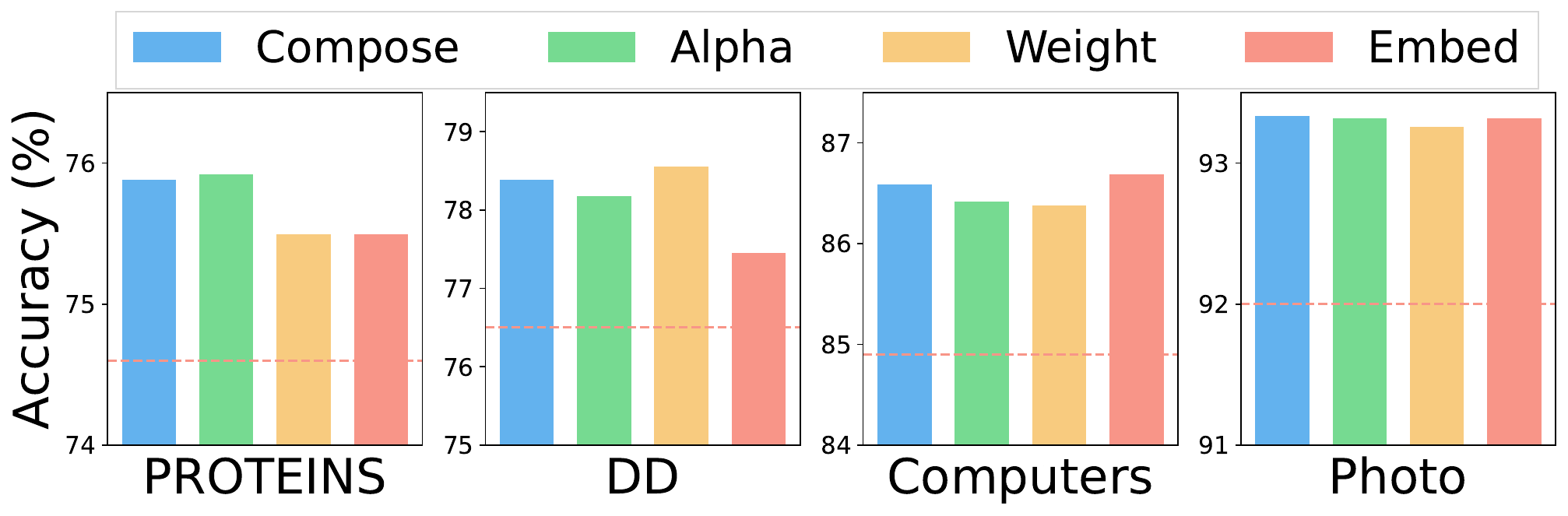}
    \label{fig:augs}
    }
    \caption{(a) Comparisons of different ablated variants of \model on real-world datasets under unsupervised settings. The horizontal dashed line refers to the results of the best-performed GNAS baseline. (b) Comparisons of different architecture augmentations of \model on real-world datasets under unsupervised settings, where `Alpha', `Weight' and `Embed' denote the augmentations from perspective of operation choices, weight and embeddings. `Compose' denotes uniformly choosing one of the three augmentations. The horizontal dashed line refers to the results of the best-performed GNAS baseline. (Best viewed in color)}
\end{figure}

\paragraph{Ablation Studies} We evaluate the effectiveness of each module of our framework by comparing the following ablated versions of our method: \smodelnosp-CONT removes our proposed contrastive search module and search architectures with the vanilla self-supervised loss. \smodelnosp-FA further replaces our proposed factor-aware training module with the vanilla self-supervised training. \smodelnosp-DISEN further replaces our proposed disentangled super-network with the vanilla super-network.

We compare the performance of the ablated versions and the best GNAS baselines on the real-world datasets under unsupervised settings. From Figure \ref{fig:ablation}, we have the following observations. First, our proposed \model outperforms all the variants on all datasets, demonstrating the effectiveness of each component of our proposed framework in searching graph architectures under unsupervised settings. Second, \smodelnosp-CONT drops drastically in performance on all datasets compared to the full version, showing the superiority of our proposed disentangled contrastive architecture search module in searching architectures. Third, the performance also decays for \smodelnosp-FA and \smodelnosp-DISEN on most datasets, showing the necessity of capturing various latent factors with different architectures. 

\begin{figure}
    \centering
    \subfloat[Computers]{\includegraphics[width=0.2\textwidth]{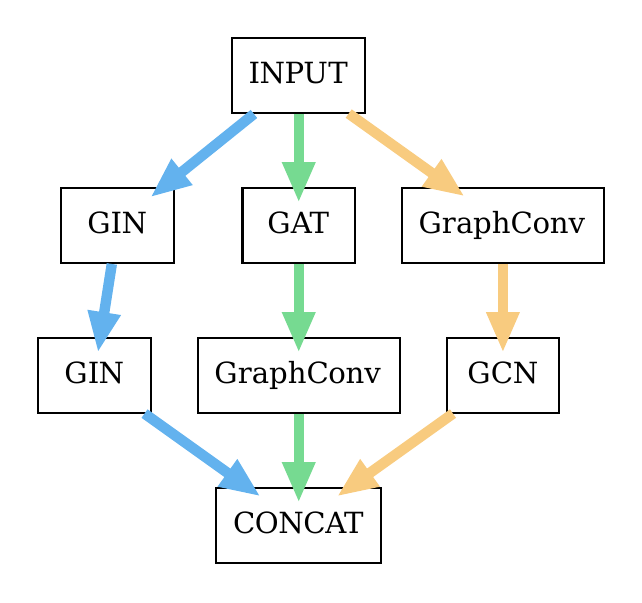}}
    \subfloat[DD]{\includegraphics[width=0.29\textwidth]{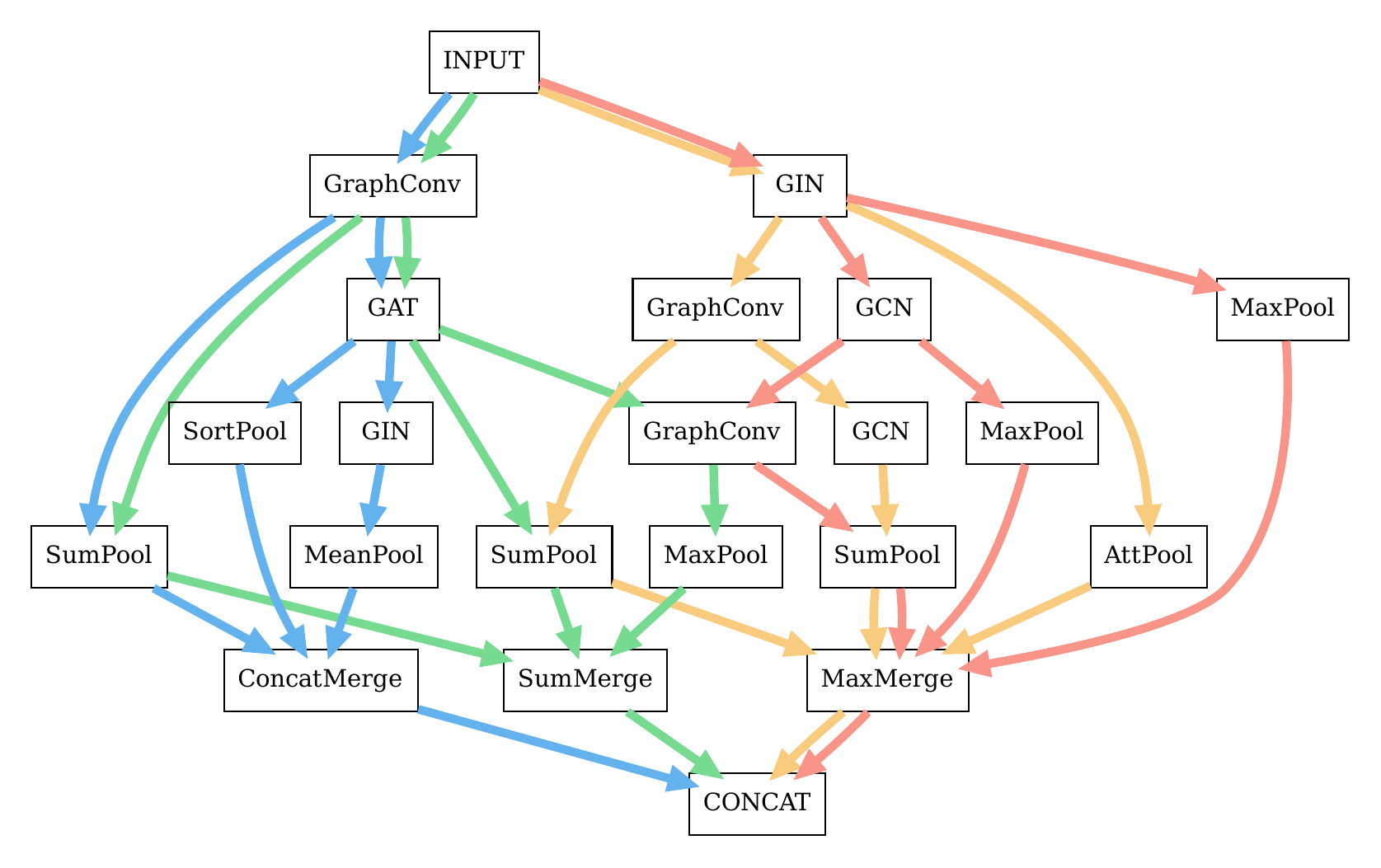}}
    \subfloat[Photo]{\includegraphics[width=0.18\textwidth]{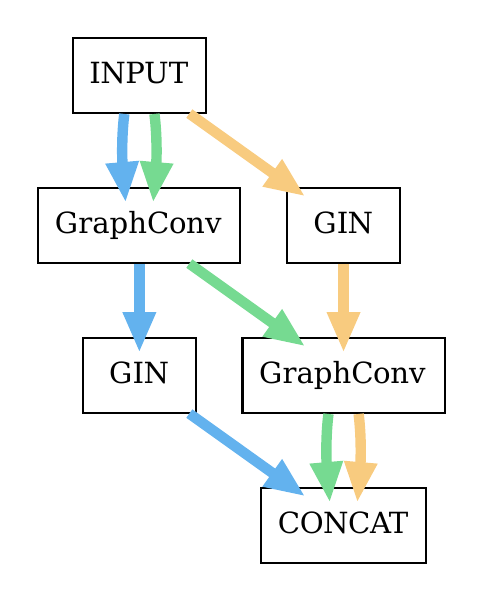}}
    \subfloat[PROTEINS]{\includegraphics[width=0.31\textwidth]{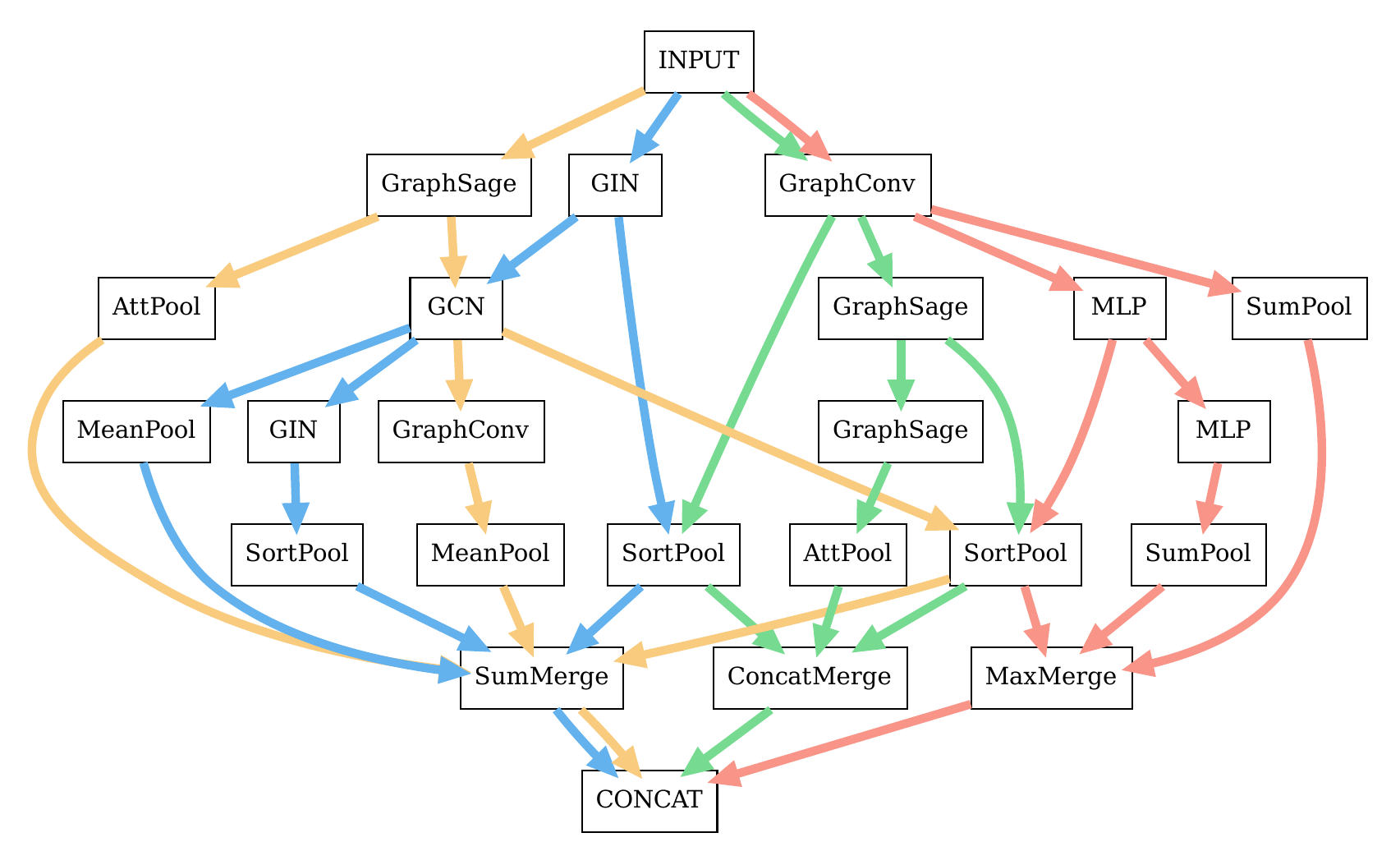}}
    \caption{Visualizations of the search architectures on different datasets. Nodes denote GNN operations except that `INPUT' denotes the input graphs with structures and features. Directed edges denote calculation flows, where different colors denote the architecture operation choices under different factors. (Best viewed in color)}
    \label{fig:visualization}

\end{figure}

\paragraph{Architecture Visualizations} We visualize the architectures searched on different unsupervised datasets in Figure ~\ref{fig:visualization}. It shows that the searched architectures of different factors adopt quite different GNN operations while sometimes sharing the same operations, which leads to an overall architecture with complex internal connections between operations. This phenomenon implies that \model can optimize the architecture operation choices as well as the operation connections for different factors to have a competitive performance on various graph datasets, which also verifies the superiority of \model in automated architecture search to save human endeavors for architecture designs.

\paragraph{Effects of Architecture Augmentations} In Figure~\ref{fig:augs}, we show the results of different architecture augmentation methods on different datasets compared with the best GNAS baseline. We find that though the best augmentations differ among datasets, they have similar performance improvements in most cases, which verifies the design of contrastive search with architecture augmentations.

\section{Related Works}

\paragraph{Graph Neural Architecture Search} Instead of manually designing more sophisticated models for various scenarios, neural architecture search, aiming to automatically discover the optimal architectures for given tasks, emerges as a hot topic recently in computer vision~\cite{ren2021comprehensive,elsken2019neural}, natural language processing~\cite{chitty2022neural}, graph representation learning~\cite{zhang2021automated,guan2021autogl,qin2022bench}, etc. In the field of graph representation learning with various applications~\cite{zhang2020deep, zhang2023large, zhang2022learning,zhang2021revisiting, zhang2023LLM4DyG}, graph neural architecture search (GNAS) methods, as the most related to our works, can be roughly classified into reinforcement-learning-based methods~\cite{gao2020graph,zhou2019auto}, evolutionary-based methods~\cite{nunes2020neural,li2020autograph,shi2020evolutionary,zhang2022deep}, and differentiable methods~\cite{ding2021diffmg,huan2021search,li2021one,cai2021rethinking,qin2021graph,qin2022graph,wei2021pooling,zhang2023dynamic,qin2023multi}. 
However, supervised labels are indispensable for the existing GNAS methods to conduct neural architecture search, which limits their applications in widely-existed scenarios where labels are scarce or not available. 

\paragraph{Unsupervised Neural Architecture Search} In unsupervised settings, some neural architecture search methods replace supervised labels with self-supervised loss during searching~\cite{kaplan2020self,timofeev2021self,nguyen2021csnas,li2021bossnas,li2022towards,hou2021automated}.
Another classic of related methods design special metrics, whose calculation does not depend on labels, as proxies for model performance estimation~\cite{zhang2021neural,mellor2021neural}. For example, UnNAS~\cite{liu2020labels} adopts pretext tasks like image rotation, coloring images, solving puzzles, etc. However, these methods are specially designed for computer vision, and can not be directly adopted to graph data. Some GNAS works exploit self-supervised loss as auxiliaries to augment the supervised search process~\cite{qin2022graph}, but the supervised labels are still mandatory for its search. 
To the best of our knowledge, this is the first work on unsupervised graph neural architecture search. 

\paragraph{Graph Self-supervised Learning} Graph self-supervised learning~\cite{liu2022graph,xie2022self} is devoted to obtaining graph representations by extracting informative knowledge with well-designed pretext tasks without labels, which can be roughly classified into contrastive~\cite{velickovic2019deep,sun2019infograph,peng2020graph,xu2021self,qiu2020gcc,han2022generative,li2022let,zhu2020deep,you2020graph,hassani2020contrastive} and generative~\cite{hou2022graphmae,li2022maskgae,xie2022self,hou2023graphmae2}, and predictive methods~\cite{you2020does,rong2020self,jin2020self}. The existing graph self-supervised learning methods usually focus on designing better pretext tasks with a fixed GNN~\cite{zhou2020graph,wu2020comprehensive,wang2019heterogeneous,wang2017community} encoder such as GCN~\cite{kipf2016semi}, GAT~\cite{velivckovic2018graph} and GIN~\cite{xu2018powerful}. Another class of related methods attempt to automate the choices of pretext tasks 
~\cite{you2021graph,yin2022autogcl,zhu2021graph,jin2021automated,ju2022multi}. We mainly consider graph neural architecture in unsupervised settings, while other pretext tasks are orthogonal to our framework and can be incorporated. 

\paragraph{Disentangled Representation Learning} 
The primary objective of disentangled representation learning is to delineate and interpret the various latent factors which influence the data we encounter in an observable context, rendering each of these factors as unique vector representations~\cite{bengio2013representation, wang2023disentangled}. It has emerged to be a useful tool in various domains, including those of computer vision~\cite{hsieh2018learning, ma2018disentangled,chen2016infogan,denton2017unsupervised,tran2017disentangled,wang2023mixup}, and recommendation systems~\cite{wang2022disentangled,chen2021curriculum,wang2021multimodal,ma2020disentangled,ma2019learning,li2021intention,wang2023curriculum,zhang2023adaptive}, graph representation learning~\cite{ma2019disentangled,liu2020independence,yang2020factorizable,li2021disentangled,li2022disentangled,zhang2023ood, zhang2023spectral, zhang2022dynamic}. As the most related,  GRACES~\cite{qin2022graph} characterize the graph latent factors inside data by designing a self-supervised disentangled graph encoder, and conduct graph neural architecture search for each graph to handle graph distribution shifts, while the training and searching process still follows the supervised paradigm. In contrast, we focus on automating the GNN designs with disentangled self-supervision in this paper.

\section{Conclusions}
In this paper, we propose a novel Disentangled Self-Supervised Graph Neural Architecture Search (\modelnosp) framework to automate the GNN designs with disentangled self-supervision, which includes disentangled graph architecture super-network, self-supervised training with joint architecture-graph disentanglement and contrastive search with architecture augmentations. Extensive experiments demonstrate that our proposed method can discover architectures with capabilities of capturing various graph latent factors and significantly outperform the state-of-the-art GNAS baselines. Detailed ablation studies and analyses show the effectiveness of our method design. One limitation is that in this paper we mainly focus on homogeneous graphs, and we leave extending our method to heterogeneous graphs for further explorations. 

\section*{Acknowledgements}
This work was supported by the National Key Research and Development Program of China No. 2020AAA0106300, National Natural Science Foundation of China (No. 62222209, 62250008, 62102222, 62206149), Beijing National Research Center for Information Science and Technology under Grant No. BNR2023RC01003, BNR2023TD03006, China National Postdoctoral Program for Innovative Talents No. BX20220185, China Postdoctoral Science Foundation No. 2022M711813, and Beijing Key Lab of Networked Multimedia. All opinions, findings, conclusions, and recommendations in this paper are those of the authors and do not necessarily reflect the views of the funding agencies.

\medskip
\small
\bibliographystyle{unsrt}
\bibliography{main}

\begin{thebibliography}{100}

\bibitem{zhang2021automated}
Ziwei Zhang, Xin Wang, and Wenwu Zhu.
\newblock Automated machine learning on graphs: A survey.
\newblock {\em arXiv preprint arXiv:2103.00742}, 2021.

\bibitem{li2020autograph}
Yaoman Li and Irwin King.
\newblock Autograph: Automated graph neural network.
\newblock In {\em International Conference on Neural Information Processing}, pages 189--201, 2020.

\bibitem{gao2020graph}
Yang Gao, Hong Yang, Peng Zhang, Chuan Zhou, and Yue Hu.
\newblock Graph neural architecture search.
\newblock In {\em IJCAI}, volume~20, pages 1403--1409, 2020.

\bibitem{wei2021pooling}
Lanning Wei, Huan Zhao, Quanming Yao, and Zhiqiang He.
\newblock Pooling architecture search for graph classification.
\newblock In {\em Proceedings of the 30th ACM International Conference on Information \& Knowledge Management}, pages 2091--2100, 2021.

\bibitem{martin2017much}
Linda Martin, Melissa Hutchens, Conrad Hawkins, and Alaina Radnov.
\newblock How much do clinical trials cost.
\newblock {\em Nat Rev Drug Discov}, 16(6):381--382, 2017.

\bibitem{paul2021artificial}
Debleena Paul, Gaurav Sanap, Snehal Shenoy, Dnyaneshwar Kalyane, Kiran Kalia, and Rakesh~K Tekade.
\newblock Artificial intelligence in drug discovery and development.
\newblock {\em Drug discovery today}, 26(1):80, 2021.

\bibitem{sun202290}
Duxin Sun, Wei Gao, Hongxiang Hu, and Simon Zhou.
\newblock Why 90\% of clinical drug development fails and how to improve it?
\newblock {\em Acta Pharmaceutica Sinica B}, 2022.

\bibitem{fan2019graph}
Wenqi Fan, Yao Ma, Qing Li, Yuan He, Eric Zhao, Jiliang Tang, and Dawei Yin.
\newblock Graph neural networks for social recommendation.
\newblock In {\em The world wide web conference}, pages 417--426, 2019.

\bibitem{ma2019disentangled}
Jianxin Ma, Peng Cui, Kun Kuang, Xin Wang, and Wenwu Zhu.
\newblock Disentangled graph convolutional networks.
\newblock In {\em International conference on machine learning}, pages 4212--4221. PMLR, 2019.

\bibitem{yang2020factorizable}
Yiding Yang, Zunlei Feng, Mingli Song, and Xinchao Wang.
\newblock Factorizable graph convolutional networks.
\newblock {\em Advances in Neural Information Processing Systems}, 33:20286--20296, 2020.

\bibitem{li2022disentangled}
Ansong Li, Zhiyong Cheng, Fan Liu, Zan Gao, Weili Guan, and Yuxin Peng.
\newblock Disentangled graph neural networks for session-based recommendation.
\newblock {\em arXiv preprint arXiv:2201.03482}, 2022.

\bibitem{cooray2022graph}
Thilini Cooray and Ngai-Man Cheung.
\newblock Graph-wise common latent factor extraction for unsupervised graph representation learning.
\newblock In {\em Proceedings of the AAAI Conference on Artificial Intelligence}, pages 6420--6428, 2022.

\bibitem{xiao2022decoupled}
Teng Xiao, Zhengyu Chen, Zhimeng Guo, Zeyang Zhuang, and Suhang Wang.
\newblock Decoupled self-supervised learning for non-homophilous graphs.
\newblock {\em arXiv preprint arXiv:2206.03601}, 2022.

\bibitem{li2021disentangled}
Haoyang Li, Xin Wang, Ziwei Zhang, Zehuan Yuan, Hang Li, and Wenwu Zhu.
\newblock Disentangled contrastive learning on graphs.
\newblock {\em Advances in Neural Information Processing Systems}, 34:21872--21884, 2021.

\bibitem{ying2018hierarchical}
Zhitao Ying, Jiaxuan You, Christopher Morris, Xiang Ren, Will Hamilton, and Jure Leskovec.
\newblock Hierarchical graph representation learning with differentiable pooling.
\newblock {\em Advances in neural information processing systems}, 31, 2018.

\bibitem{ying2019gnnexplainer}
Zhitao Ying, Dylan Bourgeois, Jiaxuan You, Marinka Zitnik, and Jure Leskovec.
\newblock Gnnexplainer: Generating explanations for graph neural networks.
\newblock {\em Advances in neural information processing systems}, 32, 2019.

\bibitem{chu2020fair}
Xiangxiang Chu, Tianbao Zhou, Bo~Zhang, and Jixiang Li.
\newblock Fair darts: Eliminating unfair advantages in differentiable architecture search.
\newblock In {\em European conference on computer vision}, pages 465--480. Springer, 2020.

\bibitem{chu2021fairnas}
Xiangxiang Chu, Bo~Zhang, and Ruijun Xu.
\newblock Fairnas: Rethinking evaluation fairness of weight sharing neural architecture search.
\newblock In {\em Proceedings of the IEEE/CVF International Conference on Computer Vision}, pages 12239--12248, 2021.

\bibitem{cha2022supernet}
Stephen Cha, Taehyeon Kim, Hayeon Lee, and Se-Young Yun.
\newblock Supernet in neural architecture search: A taxonomic survey.
\newblock {\em arXiv preprint arXiv:2204.03916}, 2022.

\bibitem{kipf2016semi}
Thomas~N Kipf and Max Welling.
\newblock Semi-supervised classification with graph convolutional networks.
\newblock {\em arXiv preprint arXiv:1609.02907}, 2016.

\bibitem{velivckovic2018graph}
Petar Veli{\v{c}}kovi{\'c}, Guillem Cucurull, Arantxa Casanova, Adriana Romero, Pietro Lio, and Yoshua Bengio.
\newblock Graph attention networks.
\newblock In {\em International Conference on Learning Representations}, 2018.

\bibitem{xu2018powerful}
Keyulu Xu, Weihua Hu, Jure Leskovec, and Stefanie Jegelka.
\newblock How powerful are graph neural networks?
\newblock {\em arXiv preprint arXiv:1810.00826}, 2018.

\bibitem{elsken2019neural}
Thomas Elsken, Jan~Hendrik Metzen, and Frank Hutter.
\newblock Neural architecture search: A survey.
\newblock {\em The Journal of Machine Learning Research}, 20(1):1997--2017, 2019.

\bibitem{pham2018efficient}
Hieu Pham, Melody Guan, Barret Zoph, Quoc Le, and Jeff Dean.
\newblock Efficient neural architecture search via parameters sharing.
\newblock In {\em International conference on machine learning}, pages 4095--4104. PMLR, 2018.

\bibitem{liu2018darts}
Hanxiao Liu, Karen Simonyan, and Yiming Yang.
\newblock Darts: Differentiable architecture search.
\newblock {\em arXiv preprint arXiv:1806.09055}, 2018.

\bibitem{guo2020single}
Zichao Guo, Xiangyu Zhang, Haoyuan Mu, Wen Heng, Zechun Liu, Yichen Wei, and Jian Sun.
\newblock Single path one-shot neural architecture search with uniform sampling.
\newblock In {\em European conference on computer vision}, pages 544--560. Springer, 2020.

\bibitem{wu2021self}
Lirong Wu, Haitao Lin, Cheng Tan, Zhangyang Gao, and Stan~Z Li.
\newblock Self-supervised learning on graphs: Contrastive, generative, or predictive.
\newblock {\em IEEE Transactions on Knowledge and Data Engineering}, 2021.

\bibitem{yan2020does}
Shen Yan, Yu~Zheng, Wei Ao, Xiao Zeng, and Mi~Zhang.
\newblock Does unsupervised architecture representation learning help neural architecture search?
\newblock {\em Advances in Neural Information Processing Systems}, 33:12486--12498, 2020.

\bibitem{luo2018neural}
Renqian Luo, Fei Tian, Tao Qin, Enhong Chen, and Tie-Yan Liu.
\newblock Neural architecture optimization.
\newblock {\em Advances in neural information processing systems}, 31, 2018.

\bibitem{ying2019bench}
Chris Ying, Aaron Klein, Eric Christiansen, Esteban Real, Kevin Murphy, and Frank Hutter.
\newblock Nas-bench-101: Towards reproducible neural architecture search.
\newblock In {\em International Conference on Machine Learning}, pages 7105--7114. PMLR, 2019.

\bibitem{zela2022surrogate}
Arber Zela, Julien~Niklas Siems, Lucas Zimmer, Jovita Lukasik, Margret Keuper, and Frank Hutter.
\newblock Surrogate nas benchmarks: Going beyond the limited search spaces of tabular nas benchmarks.
\newblock In {\em Tenth International Conference on Learning Representations}, pages 1--36. OpenReview. net, 2022.

\bibitem{you2020design}
Jiaxuan You, Zhitao Ying, and Jure Leskovec.
\newblock Design space for graph neural networks.
\newblock {\em Advances in Neural Information Processing Systems}, 33:17009--17021, 2020.

\bibitem{liu2021self}
Xiao Liu, Fanjin Zhang, Zhenyu Hou, Li~Mian, Zhaoyu Wang, Jing Zhang, and Jie Tang.
\newblock Self-supervised learning: Generative or contrastive.
\newblock {\em IEEE Transactions on Knowledge and Data Engineering}, 2021.

\bibitem{wang2022chaos}
Yifei Wang, Qi~Zhang, Yisen Wang, Jiansheng Yang, and Zhouchen Lin.
\newblock Chaos is a ladder: A new theoretical understanding of contrastive learning via augmentation overlap.
\newblock {\em arXiv preprint arXiv:2203.13457}, 2022.

\bibitem{hamilton2017inductive}
Will Hamilton, Zhitao Ying, and Jure Leskovec.
\newblock Inductive representation learning on large graphs.
\newblock {\em Advances in neural information processing systems}, 30, 2017.

\bibitem{morris2019weisfeiler}
Christopher Morris, Martin Ritzert, Matthias Fey, William~L Hamilton, Jan~Eric Lenssen, Gaurav Rattan, and Martin Grohe.
\newblock Weisfeiler and leman go neural: Higher-order graph neural networks.
\newblock In {\em Proceedings of the AAAI conference on artificial intelligence}, pages 4602--4609, 2019.

\bibitem{qin2021graph}
Yijian Qin, Xin Wang, Zeyang Zhang, and Wenwu Zhu.
\newblock Graph differentiable architecture search with structure learning.
\newblock {\em Advances in Neural Information Processing Systems}, 34, 2021.

\bibitem{dobson2003distinguishing}
Paul~D Dobson and Andrew~J Doig.
\newblock Distinguishing enzyme structures from non-enzymes without alignments.
\newblock {\em Journal of molecular biology}, 330(4):771--783, 2003.

\bibitem{shervashidze2011weisfeiler}
Nino Shervashidze, Pascal Schweitzer, Erik~Jan Van~Leeuwen, Kurt Mehlhorn, and Karsten~M Borgwardt.
\newblock Weisfeiler-lehman graph kernels.
\newblock {\em Journal of Machine Learning Research}, 12(9), 2011.

\bibitem{debnath1991structure}
Asim~Kumar Debnath, Rosa~L Lopez~de Compadre, Gargi Debnath, Alan~J Shusterman, and Corwin Hansch.
\newblock Structure-activity relationship of mutagenic aromatic and heteroaromatic nitro compounds. correlation with molecular orbital energies and hydrophobicity.
\newblock {\em Journal of medicinal chemistry}, 34(2):786--797, 1991.

\bibitem{yanardag2015deep}
Pinar Yanardag and SVN Vishwanathan.
\newblock Deep graph kernels.
\newblock In {\em Proceedings of the 21th ACM SIGKDD international conference on knowledge discovery and data mining}, pages 1365--1374, 2015.

\bibitem{morris2020tudataset}
Christopher Morris, Nils~M Kriege, Franka Bause, Kristian Kersting, Petra Mutzel, and Marion Neumann.
\newblock Tudataset: A collection of benchmark datasets for learning with graphs.
\newblock {\em arXiv preprint arXiv:2007.08663}, 2020.

\bibitem{sinha2015overview}
Arnab Sinha, Zhihong Shen, Yang Song, Hao Ma, Darrin Eide, Bo-June Hsu, and Kuansan Wang.
\newblock An overview of microsoft academic service (mas) and applications.
\newblock In {\em Proceedings of the 24th international conference on world wide web}, pages 243--246, 2015.

\bibitem{mcauley2015image}
Julian McAuley, Christopher Targett, Qinfeng Shi, and Anton Van Den~Hengel.
\newblock Image-based recommendations on styles and substitutes.
\newblock In {\em Proceedings of the 38th international ACM SIGIR conference on research and development in information retrieval}, pages 43--52, 2015.

\bibitem{hu2020open}
Weihua Hu, Matthias Fey, Marinka Zitnik, Yuxiao Dong, Hongyu Ren, Bowen Liu, Michele Catasta, and Jure Leskovec.
\newblock Open graph benchmark: Datasets for machine learning on graphs.
\newblock {\em Advances in neural information processing systems}, 33:22118--22133, 2020.

\bibitem{ren2021comprehensive}
Pengzhen Ren, Yun Xiao, Xiaojun Chang, Po-Yao Huang, Zhihui Li, Xiaojiang Chen, and Xin Wang.
\newblock A comprehensive survey of neural architecture search: Challenges and solutions.
\newblock {\em ACM Computing Surveys (CSUR)}, 54(4):1--34, 2021.

\bibitem{chitty2022neural}
Krishna~Teja Chitty-Venkata, Murali Emani, Venkatram Vishwanath, and Arun~K Somani.
\newblock Neural architecture search for transformers: A survey.
\newblock {\em IEEE Access}, 10:108374--108412, 2022.

\bibitem{guan2021autogl}
Chaoyu Guan, Ziwei Zhang, Haoyang Li, Heng Chang, Zeyang Zhang, Yijian Qin, Jiyan Jiang, Xin Wang, and Wenwu Zhu.
\newblock Autogl: A library for automated graph learning.
\newblock In {\em ICLR 2021 Workshop GTRL}, 2021.

\bibitem{qin2022bench}
Yijian Qin, Ziwei Zhang, Xin Wang, Zeyang Zhang, and Wenwu Zhu.
\newblock Nas-bench-graph: Benchmarking graph neural architecture search.
\newblock {\em arXiv preprint arXiv:2206.09166}, 2022.

\bibitem{zhang2020deep}
Ziwei Zhang, Peng Cui, and Wenwu Zhu.
\newblock Deep learning on graphs: A survey.
\newblock {\em IEEE Transactions on Knowledge and Data Engineering}, 34(1):249--270, 2020.

\bibitem{zhang2023large}
Ziwei Zhang, Haoyang Li, Zeyang Zhang, Yijian Qin, Xin Wang, and Wenwu Zhu.
\newblock Large graph models: A perspective.
\newblock {\em arXiv preprint arXiv:2308.14522}, 2023.

\bibitem{zhang2022learning}
Zeyang Zhang, Ziwei Zhang, Xin Wang, and Wenwu Zhu.
\newblock Learning to solve travelling salesman problem with hardness-adaptive curriculum.
\newblock In {\em Thirty-Fifth {AAAI} Conference on Artificial Intelligence}, 2022.

\bibitem{zhang2021revisiting}
Ziwei Zhang, Xin Wang, Zeyang Zhang, Peng Cui, and Wenwu Zhu.
\newblock Revisiting transformation invariant geometric deep learning: Are initial representations all you need?
\newblock {\em arXiv preprint arXiv:2112.12345}, 2021.

\bibitem{zhang2023LLM4DyG}
Zeyang Zhang, Xin Wang, Ziwei Zhang, Haoyang Li, Yijian Qin, Simin Wu, and Wenwu Zhu.
\newblock Llm4dyg: Can large language models solve problems on dynamic graphs?
\newblock {\em arXiv preprint}, 2023.

\bibitem{zhou2019auto}
Kaixiong Zhou, Qingquan Song, Xiao Huang, and Xia Hu.
\newblock Auto-gnn: Neural architecture search of graph neural networks.
\newblock {\em arXiv:1909.03184}, 2019.

\bibitem{nunes2020neural}
Matheus Nunes and Gisele~L Pappa.
\newblock Neural architecture search in graph neural networks.
\newblock In {\em Brazilian Conference on Intelligent Systems}, pages 302--317, 2020.

\bibitem{shi2020evolutionary}
Min Shi, David~A Wilson, Xingquan Zhu, Yu~Huang, Yuan Zhuang, Jianxun Liu, and Yufei Tang.
\newblock Genetic-gnn: Evolutionary architecture search for graph neural networks.
\newblock {\em Knowledge-Based Systems}, page 108752, 2022.

\bibitem{zhang2022deep}
Wentao Zhang, Zheyu Lin, Yu~Shen, Yang Li, Zhi Yang, and Bin Cui.
\newblock Deep and flexible graph neural architecture search.
\newblock In {\em International Conference on Machine Learning}, pages 26362--26374. PMLR, 2022.

\bibitem{ding2021diffmg}
Yuhui Ding, Quanming Yao, Huan Zhao, and Tong Zhang.
\newblock Diffmg: Differentiable meta graph search for heterogeneous graph neural networks.
\newblock In {\em Proceedings of the 27th ACM SIGKDD Conference on Knowledge Discovery \& Data Mining}, pages 279--288, 2021.

\bibitem{huan2021search}
ZHAO Huan, YAO Quanming, and TU~Weiwei.
\newblock Search to aggregate neighborhood for graph neural network.
\newblock In {\em 2021 IEEE 37th International Conference on Data Engineering}, pages 552--563, 2021.

\bibitem{li2021one}
Yanxi Li, Zean Wen, Yunhe Wang, and Chang Xu.
\newblock One-shot graph neural architecture search with dynamic search space.
\newblock In {\em Proceedings of the AAAI Conference on Artificial Intelligence}, volume~35, pages 8510--8517, 2021.

\bibitem{cai2021rethinking}
Shaofei Cai, Liang Li, Jincan Deng, Beichen Zhang, Zheng-Jun Zha, Li~Su, and Qingming Huang.
\newblock Rethinking graph neural architecture search from message-passing.
\newblock In {\em Proceedings of the IEEE/CVF Conference on Computer Vision and Pattern Recognition}, pages 6657--6666, 2021.

\bibitem{qin2022graph}
Yijian Qin, Xin Wang, Ziwei Zhang, Pengtao Xie, and Wenwu Zhu.
\newblock Graph neural architecture search under distribution shifts.
\newblock In {\em International Conference on Machine Learning}, pages 18083--18095. PMLR, 2022.

\bibitem{zhang2023dynamic}
Zeyang Zhang, Ziwei Zhang, Xin Wang, Yijian Qin, Zhou Qin, and Wenwu Zhu.
\newblock Dynamic heterogeneous graph attention neural architecture search.
\newblock In {\em Thirty-Seventh {AAAI} Conference on Artificial Intelligence}, 2023.

\bibitem{qin2023multi}
Yijian Qin, Xin Wang, Ziwei Zhang, Hong Chen, and Wenwu Zhu.
\newblock Multi-task graph neural architecture search with task-aware collaboration and curriculum.
\newblock {\em Advances in neural information processing systems}, 2023.

\bibitem{kaplan2020self}
Sapir Kaplan and Raja Giryes.
\newblock Self-supervised neural architecture search.
\newblock {\em arXiv preprint arXiv:2007.01500}, 2020.

\bibitem{timofeev2021self}
Aleksandr Timofeev, Grigorios~G Chrysos, and Volkan Cevher.
\newblock Self-supervised neural architecture search for imbalanced datasets.
\newblock {\em arXiv preprint arXiv:2109.08580}, 2021.

\bibitem{nguyen2021csnas}
Nam Nguyen and J~Morris Chang.
\newblock Csnas: Contrastive self-supervised learning neural architecture search via sequential model-based optimization.
\newblock {\em IEEE Transactions on Artificial Intelligence}, 3(4):609--624, 2021.

\bibitem{li2021bossnas}
Changlin Li, Tao Tang, Guangrun Wang, Jiefeng Peng, Bing Wang, Xiaodan Liang, and Xiaojun Chang.
\newblock Bossnas: Exploring hybrid cnn-transformers with block-wisely self-supervised neural architecture search.
\newblock In {\em Proceedings of the IEEE/CVF International Conference on Computer Vision}, pages 12281--12291, 2021.

\bibitem{li2022towards}
Zhuowei Li, Yibo Gao, Zhenzhou Zha, Zhiqiang Hu, Qing Xia, Shaoting Zhang, and Dimitris~N Metaxas.
\newblock Towards self-supervised and weight-preserving neural architecture search.
\newblock {\em arXiv preprint arXiv:2206.04125}, 2022.

\bibitem{hou2021automated}
Zhenyu Hou, Yukuo Cen, Yuxiao Dong, Jie Zhang, and Jie Tang.
\newblock Automated unsupervised graph representation learning.
\newblock {\em IEEE Transactions on Knowledge and Data Engineering}, 2021.

\bibitem{zhang2021neural}
Xuanyang Zhang, Pengfei Hou, Xiangyu Zhang, and Jian Sun.
\newblock Neural architecture search with random labels.
\newblock In {\em Proceedings of the IEEE/CVF Conference on Computer Vision and Pattern Recognition}, pages 10907--10916, 2021.

\bibitem{mellor2021neural}
Joe Mellor, Jack Turner, Amos Storkey, and Elliot~J Crowley.
\newblock Neural architecture search without training.
\newblock In {\em International Conference on Machine Learning}, pages 7588--7598. PMLR, 2021.

\bibitem{liu2020labels}
Chenxi Liu, Piotr Doll{\'a}r, Kaiming He, Ross Girshick, Alan Yuille, and Saining Xie.
\newblock Are labels necessary for neural architecture search?
\newblock In {\em European Conference on Computer Vision}, pages 798--813. Springer, 2020.

\bibitem{liu2022graph}
Yixin Liu, Ming Jin, Shirui Pan, Chuan Zhou, Yu~Zheng, Feng Xia, and Philip Yu.
\newblock Graph self-supervised learning: A survey.
\newblock {\em IEEE Transactions on Knowledge and Data Engineering}, 2022.

\bibitem{xie2022self}
Yaochen Xie, Zhao Xu, and Shuiwang Ji.
\newblock Self-supervised representation learning via latent graph prediction.
\newblock {\em arXiv preprint arXiv:2202.08333}, 2022.

\bibitem{velickovic2019deep}
Petar Velickovic, William Fedus, William~L Hamilton, Pietro Li{\`o}, Yoshua Bengio, and R~Devon Hjelm.
\newblock Deep graph infomax.
\newblock {\em ICLR (Poster)}, 2(3):4, 2019.

\bibitem{sun2019infograph}
Fan-Yun Sun, Jordan Hoffmann, Vikas Verma, and Jian Tang.
\newblock Infograph: Unsupervised and semi-supervised graph-level representation learning via mutual information maximization.
\newblock {\em arXiv preprint arXiv:1908.01000}, 2019.

\bibitem{peng2020graph}
Zhen Peng, Wenbing Huang, Minnan Luo, Qinghua Zheng, Yu~Rong, Tingyang Xu, and Junzhou Huang.
\newblock Graph representation learning via graphical mutual information maximization.
\newblock In {\em Proceedings of The Web Conference 2020}, pages 259--270, 2020.

\bibitem{xu2021self}
Minghao Xu, Hang Wang, Bingbing Ni, Hongyu Guo, and Jian Tang.
\newblock Self-supervised graph-level representation learning with local and global structure.
\newblock In {\em International Conference on Machine Learning}, pages 11548--11558. PMLR, 2021.

\bibitem{qiu2020gcc}
Jiezhong Qiu, Qibin Chen, Yuxiao Dong, Jing Zhang, Hongxia Yang, Ming Ding, Kuansan Wang, and Jie Tang.
\newblock Gcc: Graph contrastive coding for graph neural network pre-training.
\newblock In {\em Proceedings of the 26th ACM SIGKDD International Conference on Knowledge Discovery \& Data Mining}, pages 1150--1160, 2020.

\bibitem{han2022generative}
Yuehui Han, Le~Hui, Haobo Jiang, Jianjun Qian, and Jin Xie.
\newblock Generative subgraph contrast for self-supervised graph representation learning.
\newblock {\em arXiv preprint arXiv:2207.11996}, 2022.

\bibitem{li2022let}
Sihang Li, Xiang Wang, An~Zhang, Yingxin Wu, Xiangnan He, and Tat-Seng Chua.
\newblock Let invariant rationale discovery inspire graph contrastive learning.
\newblock In {\em International Conference on Machine Learning}, pages 13052--13065. PMLR, 2022.

\bibitem{zhu2020deep}
Yanqiao Zhu, Yichen Xu, Feng Yu, Qiang Liu, Shu Wu, and Liang Wang.
\newblock Deep graph contrastive representation learning.
\newblock {\em arXiv preprint arXiv:2006.04131}, 2020.

\bibitem{you2020graph}
Yuning You, Tianlong Chen, Yongduo Sui, Ting Chen, Zhangyang Wang, and Yang Shen.
\newblock Graph contrastive learning with augmentations.
\newblock {\em Advances in Neural Information Processing Systems}, 33:5812--5823, 2020.

\bibitem{hassani2020contrastive}
Kaveh Hassani and Amir~Hosein Khasahmadi.
\newblock Contrastive multi-view representation learning on graphs.
\newblock In {\em International Conference on Machine Learning}, pages 4116--4126. PMLR, 2020.

\bibitem{hou2022graphmae}
Zhenyu Hou, Xiao Liu, Yuxiao Dong, Chunjie Wang, Jie Tang, et~al.
\newblock Graphmae: Self-supervised masked graph autoencoders.
\newblock {\em arXiv preprint arXiv:2205.10803}, 2022.

\bibitem{li2022maskgae}
Jintang Li, Ruofan Wu, Wangbin Sun, Liang Chen, Sheng Tian, Liang Zhu, Changhua Meng, Zibin Zheng, and Weiqiang Wang.
\newblock Maskgae: Masked graph modeling meets graph autoencoders.
\newblock {\em arXiv preprint arXiv:2205.10053}, 2022.

\bibitem{hou2023graphmae2}
Zhenyu Hou, Yufei He, Yukuo Cen, Xiao Liu, Yuxiao Dong, Evgeny Kharlamov, and Jie Tang.
\newblock Graphmae2: A decoding-enhanced masked self-supervised graph learner.
\newblock In {\em Proceedings of the ACM Web Conference 2023}, pages 737--746, 2023.

\bibitem{you2020does}
Yuning You, Tianlong Chen, Zhangyang Wang, and Yang Shen.
\newblock When does self-supervision help graph convolutional networks?
\newblock In {\em international conference on machine learning}, pages 10871--10880. PMLR, 2020.

\bibitem{rong2020self}
Yu~Rong, Yatao Bian, Tingyang Xu, Weiyang Xie, Ying Wei, Wenbing Huang, and Junzhou Huang.
\newblock Self-supervised graph transformer on large-scale molecular data.
\newblock {\em Advances in Neural Information Processing Systems}, 33:12559--12571, 2020.

\bibitem{jin2020self}
Wei Jin, Tyler Derr, Haochen Liu, Yiqi Wang, Suhang Wang, Zitao Liu, and Jiliang Tang.
\newblock Self-supervised learning on graphs: Deep insights and new direction.
\newblock {\em arXiv preprint arXiv:2006.10141}, 2020.

\bibitem{zhou2020graph}
Jie Zhou, Ganqu Cui, Shengding Hu, Zhengyan Zhang, Cheng Yang, Zhiyuan Liu, Lifeng Wang, Changcheng Li, and Maosong Sun.
\newblock Graph neural networks: A review of methods and applications.
\newblock {\em AI open}, 1:57--81, 2020.

\bibitem{wu2020comprehensive}
Zonghan Wu, Shirui Pan, Fengwen Chen, Guodong Long, Chengqi Zhang, and S~Yu Philip.
\newblock A comprehensive survey on graph neural networks.
\newblock {\em IEEE transactions on neural networks and learning systems}, 32(1):4--24, 2020.

\bibitem{wang2019heterogeneous}
Xiao Wang, Houye Ji, Chuan Shi, Bai Wang, Yanfang Ye, Peng Cui, and Philip~S Yu.
\newblock Heterogeneous graph attention network.
\newblock In {\em The world wide web conference}, pages 2022--2032, 2019.

\bibitem{wang2017community}
Xiao Wang, Peng Cui, Jing Wang, Jian Pei, Wenwu Zhu, and Shiqiang Yang.
\newblock Community preserving network embedding.
\newblock In {\em Proceedings of the AAAI conference on artificial intelligence}, volume~31, 2017.

\bibitem{you2021graph}
Yuning You, Tianlong Chen, Yang Shen, and Zhangyang Wang.
\newblock Graph contrastive learning automated.
\newblock In {\em International Conference on Machine Learning}, pages 12121--12132. PMLR, 2021.

\bibitem{yin2022autogcl}
Yihang Yin, Qingzhong Wang, Siyu Huang, Haoyi Xiong, and Xiang Zhang.
\newblock Autogcl: Automated graph contrastive learning via learnable view generators.
\newblock In {\em Proceedings of the AAAI Conference on Artificial Intelligence}, pages 8892--8900, 2022.

\bibitem{zhu2021graph}
Yanqiao Zhu, Yichen Xu, Feng Yu, Qiang Liu, Shu Wu, and Liang Wang.
\newblock Graph contrastive learning with adaptive augmentation.
\newblock In {\em Proceedings of the Web Conference 2021}, pages 2069--2080, 2021.

\bibitem{jin2021automated}
Wei Jin, Xiaorui Liu, Xiangyu Zhao, Yao Ma, Neil Shah, and Jiliang Tang.
\newblock Automated self-supervised learning for graphs.
\newblock {\em arXiv preprint arXiv:2106.05470}, 2021.

\bibitem{ju2022multi}
Mingxuan Ju, Tong Zhao, Qianlong Wen, Wenhao Yu, Neil Shah, Yanfang Ye, and Chuxu Zhang.
\newblock Multi-task self-supervised graph neural networks enable stronger task generalization.
\newblock {\em arXiv preprint arXiv:2210.02016}, 2022.

\bibitem{bengio2013representation}
Yoshua Bengio, Aaron Courville, and Pascal Vincent.
\newblock Representation learning: A review and new perspectives.
\newblock {\em IEEE transactions on pattern analysis and machine intelligence}, 35(8):1798--1828, 2013.

\bibitem{wang2023disentangled}
Xin Wang, Hong Chen, Si'ao Tang, Zihao Wu, and Wenwu Zhu.
\newblock Disentangled representation learning, 2023.

\bibitem{hsieh2018learning}
Jun-Ting Hsieh, Bingbin Liu, De-An Huang, Li~F Fei-Fei, and Juan~Carlos Niebles.
\newblock Learning to decompose and disentangle representations for video prediction.
\newblock {\em Advances in neural information processing systems}, 31, 2018.

\bibitem{ma2018disentangled}
Liqian Ma, Qianru Sun, Stamatios Georgoulis, Luc Van~Gool, Bernt Schiele, and Mario Fritz.
\newblock Disentangled person image generation.
\newblock In {\em Proceedings of the IEEE Conference on Computer Vision and Pattern Recognition}, pages 99--108, 2018.

\bibitem{chen2016infogan}
Xi~Chen, Yan Duan, Rein Houthooft, John Schulman, Ilya Sutskever, and Pieter Abbeel.
\newblock Infogan: Interpretable representation learning by information maximizing generative adversarial nets.
\newblock {\em Advances in neural information processing systems}, 29, 2016.

\bibitem{denton2017unsupervised}
Emily~L Denton et~al.
\newblock Unsupervised learning of disentangled representations from video.
\newblock {\em Advances in neural information processing systems}, 30, 2017.

\bibitem{tran2017disentangled}
Luan Tran, Xi~Yin, and Xiaoming Liu.
\newblock Disentangled representation learning gan for pose-invariant face recognition.
\newblock In {\em Proceedings of the IEEE conference on computer vision and pattern recognition}, pages 1415--1424, 2017.

\bibitem{wang2023mixup}
Xin Wang, Zihao Wu, Hong Chen, Xiaohan Lan, and Wenwu Zhu.
\newblock Mixup-augmented temporally debiased video grounding with content-location disentanglement.
\newblock 2023.

\bibitem{wang2022disentangled}
Xin Wang, Hong Chen, Yuwei Zhou, Jianxin Ma, and Wenwu Zhu.
\newblock Disentangled representation learning for recommendation.
\newblock {\em IEEE Transactions on Pattern Analysis and Machine Intelligence}, 2022.

\bibitem{chen2021curriculum}
Hong Chen, Yudong Chen, Xin Wang, Ruobing Xie, Rui Wang, Feng Xia, and Wenwu Zhu.
\newblock Curriculum disentangled recommendation with noisy multi-feedback.
\newblock {\em Advances in Neural Information Processing Systems}, 34:26924--26936, 2021.

\bibitem{wang2021multimodal}
Xin Wang, Hong Chen, and Wenwu Zhu.
\newblock Multimodal disentangled representation for recommendation.
\newblock In {\em 2021 IEEE International Conference on Multimedia and Expo (ICME)}, pages 1--6, 2021.

\bibitem{ma2020disentangled}
Jianxin Ma, Chang Zhou, Hongxia Yang, Peng Cui, Xin Wang, and Wenwu Zhu.
\newblock Disentangled self-supervision in sequential recommenders.
\newblock In {\em Proceedings of the 26th ACM SIGKDD International Conference on Knowledge Discovery \& Data Mining}, pages 483--491, 2020.

\bibitem{ma2019learning}
Jianxin Ma, Chang Zhou, Peng Cui, Hongxia Yang, and Wenwu Zhu.
\newblock Learning disentangled representations for recommendation.
\newblock {\em Advances in neural information processing systems}, 32, 2019.

\bibitem{li2021intention}
Haoyang Li, Xin Wang, Ziwei Zhang, Jianxin Ma, Peng Cui, and Wenwu Zhu.
\newblock Intention-aware sequential recommendation with structured intent transition.
\newblock {\em IEEE Transactions on Knowledge and Data Engineering}, 2021.

\bibitem{wang2023curriculum}
Xin Wang, Zirui Pan, Yuwei Zhou, Hong Chen, Chendi Ge, and Wenwu Zhu.
\newblock Curriculum co-disentangled representation learning across multiple environments for social recommendation.
\newblock 2023.

\bibitem{zhang2023adaptive}
Yipeng Zhang, Xin Wang, Hong Chen, and Wenwu Zhu.
\newblock Adaptive disentangled transformer for sequential recommendation.
\newblock In {\em Proceedings of the 29th ACM SIGKDD Conference on Knowledge Discovery and Data Mining}, pages 3434--3445, 2023.

\bibitem{liu2020independence}
Yanbei Liu, Xiao Wang, Shu Wu, and Zhitao Xiao.
\newblock Independence promoted graph disentangled networks.
\newblock In {\em Proceedings of the AAAI Conference on Artificial Intelligence}, volume~34, pages 4916--4923, 2020.

\bibitem{zhang2023ood}
Zeyang Zhang, Xingwang Li, Fei Teng, Ning Lin, Xueling Zhu, Xin Wang, and Wenwu Zhu.
\newblock Out-of-distribution generalized dynamic graph neural network for human albumin prediction.
\newblock In {\em IEEE International Conference on Medical Artificial Intelligence}, 2023.

\bibitem{zhang2023spectral}
Zeyang Zhang, Xin Wang, Ziwei Zhang, Zhou Qin, Weigao Wen, Hui Xue, Haoyang Li, and Wenwu Zhu.
\newblock Spectral invariant learning for dynamic graphs under distribution shifts.
\newblock In {\em Advances in Neural Information Processing Systems}, 2023.

\bibitem{zhang2022dynamic}
Zeyang Zhang, Xin Wang, Ziwei Zhang, Haoyang Li, Zhou Qin, and Wenwu Zhu.
\newblock Dynamic graph neural networks under spatio-temporal distribution shift.
\newblock In {\em Advances in Neural Information Processing Systems}, 2022.

\bibitem{sechidis2011stratification}
Konstantinos Sechidis, Grigorios Tsoumakas, and Ioannis Vlahavas.
\newblock On the stratification of multi-label data.
\newblock In {\em Machine Learning and Knowledge Discovery in Databases: European Conference, ECML PKDD 2011, Athens, Greece, September 5-9, 2011, Proceedings, Part III 22}, pages 145--158. Springer, 2011.

\bibitem{zhang2018end}
Muhan Zhang, Zhicheng Cui, Marion Neumann, and Yixin Chen.
\newblock An end-to-end deep learning architecture for graph classification.
\newblock In {\em Proceedings of the AAAI conference on artificial intelligence}, volume~32, 2018.

\bibitem{li2015gated}
Yujia Li, Daniel Tarlow, Marc Brockschmidt, and Richard Zemel.
\newblock Gated graph sequence neural networks.
\newblock {\em arXiv preprint arXiv:1511.05493}, 2015.

\bibitem{ioffe2015batch}
Sergey Ioffe and Christian Szegedy.
\newblock Batch normalization: Accelerating deep network training by reducing internal covariate shift.
\newblock In {\em International conference on machine learning}, pages 448--456. PMLR, 2015.

\bibitem{ba2016layer}
Jimmy~Lei Ba, Jamie~Ryan Kiros, and Geoffrey~E Hinton.
\newblock Layer normalization.
\newblock {\em arXiv preprint arXiv:1607.06450}, 2016.

\bibitem{kingma2014adam}
Diederik~P Kingma and Jimmy Ba.
\newblock Adam: A method for stochastic optimization.
\newblock {\em arXiv preprint arXiv:1412.6980}, 2014.

\bibitem{paszke2019pytorch}
Adam Paszke, Sam Gross, Francisco Massa, Adam Lerer, James Bradbury, Gregory Chanan, Trevor Killeen, Zeming Lin, Natalia Gimelshein, Luca Antiga, et~al.
\newblock Pytorch: An imperative style, high-performance deep learning library.
\newblock {\em Advances in neural information processing systems}, 32, 2019.

\bibitem{Fey/Lenssen/2019}
Matthias Fey and Jan~E. Lenssen.
\newblock Fast graph representation learning with {PyTorch Geometric}.
\newblock In {\em ICLR Workshop on Representation Learning on Graphs and Manifolds}, 2019.

\end{thebibliography}

\clearpage
\newpage
\appendix

\section{Notations}
\begin{table}[htbp]
\caption{Summary of notations and their descriptions. }
\centering
\begin{tabular}{c|l}
\toprule
\textbf{Notations}                                                  & \textbf{Descriptions}                                                                   \\ \midrule
$\mathcal{G},\mathcal{Y},\mathcal{A},\mathcal{W}$                    & Graph space, label space, architecture space and weight space                                         \\
$\alpha,w$              & Architecture operation choices and  architecture weight                                                       \\
$\mathbf{H},\mathbf{A}$                                    & Graph embeddings and adjacency matrix                                        \\
$\mathcal{O},|\mathcal{O}|$                                    & A pool of GNN operations and the number of operations  \\
$f_{\alpha,w}(\cdot)$                                    & A GNN characterized by operation choices $\alpha$ and weight $w$  \\
$s(\cdot)$                                    & Pseudo label generator defined by pretext tasks  \\
$\mathbf{c}$                                    & Learnable vectors for the latent factors  \\
$K$                                    & The number of the latent factors \\
$r$                                    & Perturbation ratio in architecture augmentations \\
$T_f(\cdot)$                                    & Architecture augmentation function\\
$\text{Enc}(\cdot)$                                    & Architecture encoding function \\
$l,\mathcal{L}$                                    & loss functions \\
$\phi(\cdot)$  & A function that calculates the similarity of two embeddings 
                                \\ \bottomrule
\end{tabular}
\end{table}

\section{Additional Experiments and Analyses}
\subsection{Complexity Analysis} 
\label{sec:complexity}
Denote the number of nodes and edges in the graph as $N$ and $E$, the number of latent factors as $K$, the number of operation choices as $|\mathcal{O}|$, the dimensionality of hidden representations as $d$. The time complexity of the disentangled super-network is $O(K|E|d + K|V|d^2)$, where the computation for each factor is fully parallelizable and amenable to GPU acceleration, and $K$ is usually a small constant. The time complexity of the self-supervised training and contrastive search modules is both $O(K^2d^2)$. As architectures under different factors share the parameters, the number of learnable parameters is the same as classical graph super-network, i.e., $O(|\mathcal{O}|d^2)$. Therefore, the complexity of our method is comparable to classical GNAS methods.

\subsection{Empirical Running Time} 
\begin{table}[]
\centering
\caption{Comparisons of NAS methods in terms of empirical running time and performance on unsupervised graph classification datasets (with single NVIDIA GeForce RTX 3090).}
\label{tab:time}
\adjustbox{max width=0.98\textwidth}{
\begin{tabular}{ccccccccc}
\toprule
\textbf{Data} & \multicolumn{2}{c}{\textbf{PROTEINS}} & \multicolumn{2}{c}{\textbf{DD}} & \multicolumn{2}{c}{\textbf{MUTAG}} & \multicolumn{2}{c}{\textbf{IMDB-B}} \\
Metric        & ACC(\%)                 & Time(s)      & ACC(\%)              & Time(s)   & ACC(\%)                & Time(s)    & ACC(\%)                & Time(s)     \\ \midrule
Random        & \ms{74.5}{0.9}         & 2952         & \ms{74.8}{1.3}      & 9401      & \ms{82.1}{2.8}        & 949        & \ms{69.0}{2.1}        & 2535        \\
DARTS         & \ms{73.6}{0.9}         & 80           & \ms{75.7}{0.9}      & 650       & \ms{86.5}{2.3}        & 21         & \ms{70.4}{0.6}        & 65          \\
GraphNAS      & \ms{73.6}{0.7}         & 1897         & \ms{75.2}{0.9}      & 7830      & \ms{77.5}{0.7}        & 273        & \ms{62.7}{1.3}        & 1595        \\
PAS           & \mstwo{74.6}{0.3}      & 156          & \ms{76.5}{0.9}      & 931       & \ms{84.0}{1.6}        & 36         & \ms{64.6}{13.8}       & 127         \\
GASSO         & -                      &              & -                   &           & -                     &            & -                     &             \\ \midrule
\model        & \msone{76.0}{0.2}      & 471          & \msone{78.4}{0.7}   & 1800      & \msone{88.7}{0.7}     & 41         & \msone{72.0}{0.5}     & 261         \\ \bottomrule
\end{tabular}
}
\end{table}

\begin{table}[]
\centering
\caption{Comparisons of NAS methods in terms of empirical running time and performance on unsupervised node classification datasets (with single NVIDIA GeForce RTX 3090).}
\label{tab:time2}
\adjustbox{max width=0.98\textwidth}{
\begin{tabular}{ccccccccc}
\toprule
\textbf{Data} & \multicolumn{2}{c}{\textbf{CS}} & \multicolumn{2}{c}{\textbf{Computers}} & \multicolumn{2}{c}{\textbf{Physics}} & \multicolumn{2}{c}{\textbf{Photo}} \\
Metric        & ACC(\%)             & Time(s)   & ACC(\%)                 & Time(s)      & ACC(\%)                & Time(s)     & ACC(\%)               & Time(s)    \\ \midrule
Random        & \ms{92.9}{0.3}      & 1071      & \ms{84.8}{0.4}          & 3605         & \ms{95.4}{0.1}         & 2095        & \ms{91.1}{0.6}        & 522        \\
DARTS         & \ms{92.8}{0.3}      & 34        & \ms{79.7}{0.5}          & 79           & \ms{95.2}{0.1}         & 75          & \ms{91.5}{0.6}        & 13         \\
GraphNAS      & \ms{91.6}{0.3}      & 647       & \ms{69.0}{0.6}          & 5295         & \ms{94.5}{0.1}         & 2268        & \ms{89.3}{0.7}        & 435        \\
PAS           & -                   &           & -                       &              & -                      &             & -                     &            \\
GASSO         & \ms{93.1}{0.3}      & 34        & \ms{84.9}{0.4}          & 69           & \mstwo{95.7}{0.1}      & 75          & \mstwo{92.0}{0.3}     & 13         \\ \midrule
\model        & \msone{93.5}{0.2}   & 49        & \msone{86.6}{0.4}       & 201          & \msone{95.7}{0.1}      & 99          & \msone{93.3}{0.3}     & 20         \\ \bottomrule
\end{tabular}
}
\end{table}

We make the comparisons of different NAS methods in terms of the empirical running time. The time is tested with one NVIDIA 3090 GPU. As shown in the Table~\ref{tab:time} and Table~\ref{tab:time2}, the running time of our method \model is on par with the state-of-the-art one-shot NAS methods (e.g., DARTS, GASSO, and PAS), which is much more efficient than the multi-trial NAS methods (e.g., random search and GraphNAS). While being competitive in efficiency, our method has significant performance improvements over the baselines. The empirical results also confirm the theoretical complexity analysis in Section \ref{sec:complexity} that our method does not introduce many additional computational costs.

\subsection{Search Space Analysis}

\begin{table}[]
\centering
\caption{The performance of our method with increasing number of available GNN options on the Computers dataset.}
\label{tab:ab_gnn}
\begin{tabular}{ccccc}
\toprule
$|\mathcal{O}|$                    & 2                    & 3                    & 4                    & 5                    \\ \midrule
ACC(\%)               & \ms{84.8}{0.4}            & \ms{86.2}{0.3}            & \ms{86.6}{0.3}            & \ms{86.6}{0.4}            \\ \bottomrule
\end{tabular}
\end{table}

\begin{table}[]
\centering
\caption{The performance of our method with increasing number of available factors on the Computers dataset.}
\label{tab:ab_factor}
\begin{tabular}{ccccc}
\toprule
$K$                    & 2                    & 3                    & 4                    & 5                    \\ \midrule
ACC(\%)               & \ms{85.1}{0.4}            & \ms{86.6}{0.4}            & \ms{87.3}{0.4}            & \ms{86.5}{0.4}            \\ \bottomrule
\end{tabular}
\end{table}

We show how the performance of \model changes when the search space is larger on the Computers dataset in the Table~\ref{tab:ab_gnn} and Table~\ref{tab:ab_factor}. In Table~\ref{tab:ab_gnn} , we enlarge the search space by gradually increasing the GNN operation pool 
, i.e. increasing the number of available GNN options. In Table~\ref{tab:ab_factor}, we enlarge the search space by gradually increasing the number of factors 
, i.e. increasing the number of paths to capture graph factors. As shown in the tables, when the search space is larger, the performance of our method gradually improves, which verifies that our method can discover better architectures with a larger search space.

\subsection{Discussions of the Searched Architectures}
We visualize several searched architectures in Figure 4 of the main paper, which are powerful yet complex. Here, we make following discussions about the searched architectures.

We observe that in different random training runs, while the searched architectures show similar performance, their DAGs are not the same, which is consistent with the NAS literature~\cite{ying2019bench, zela2022surrogate}. A possible reason is that there exist plenty of different architectures with very similar performance in the large graph architecture search space~\cite{qin2022bench}. As shown in Table 2 in the main paper, our method has relatively low performance variance and high performance expectation, which shows that our method can better search for the potential top-ranked architectures than baselines. 

The number of factors $K$, which reflects the assumption of the number of graph factors to be captured inside the data, controls the searchable architectures in our method. When $K=1$, our method can include single simple architectures with arbitrary operation combinations. When $K\geq 1$, our method can discover more sophisticated architectures to capture the inherent graph factors and obtain better performance. Empirically, we observe that when $K\geq 1$, the searched architectures are more complex than single architectures but are also more competitive in capturing the graph properties, which verifies the design of our method. 

\subsection{Additional Results in Unsupervised Settings}

\begin{table}[]
\centering
\caption{The performance of NAS methods on other graph datasets in unsupervised settings.}
\label{tab:cora}
\begin{tabular}{cccc}
\toprule
\textbf{Data} & \textbf{Cora} & \textbf{Citeseer} & \textbf{Pubmed} \\ \midrule
DARTS         & \ms{78.4}{0.3}     & \ms{71.1}{0.8}         & \ms{78.8}{0.7}       \\
GraphNAS      & \ms{81.5}{0.6}     & \ms{70.4}{1.1}         & \ms{79.9}{0.9}       \\
GASSO         & \ms{80.2}{0.8}     & \ms{69.5}{1.1}         & \ms{78.1}{0.8}       \\ \midrule
\model         & \msone{83.5}{0.4}     & \msone{72.2}{1.4}         & \msone{80.6}{0.4}       \\ \bottomrule
\end{tabular}
\end{table}

We provide the experimental results on Cora, CiteSeer, and PubMed in Table~\ref{tab:cora}. We follow the public data splits~\cite{velickovic2019deep} of Cora, Citeseer, and Pubmed, and conduct graph neural architecture search without labels. Similar to other unsupervised node classification datasets in the paper, we train the super-network with fixed epochs , and for evaluation, we train a linear classifier and report the mean accuracy and standard deviations on the test nodes of 5 runs with different random seeds. As shown in the Table~\ref{tab:cora}, our method \model has significant performance improvement over the NAS baselines.

\section{Experimental Details}
\subsection{Unsupervised Settings}
\paragraph{Setups} Following previous works of graph self-supervised learning~\cite{you2020graph,xie2022self}, we first pretrain the models by self-supervised loss with fixed epochs, and then evaluate the models by finetuning an extra classifier. As supervised labels are not available in unsupervised settings, for fair comparisons, all the methods adopt the same self-supervised tasks, ~\cite{you2020graph} and ~\cite{xie2022self} for graph and node classification tasks respectively. For GNAS baselines, the self-supervised loss is utilized to train the model parameters as well as select the architectures. 

\paragraph{Evaluation protocols} For graph-level classification tasks, the obtained graph representations are evaluated by an SVM classifier with a 10-fold cross-validation and the process is repeated by five times with different seeds. For node-level classification tasks, the obtained node representations are evaluated by a logistic regression classifier with random splits twenty times. The average accuracies and their standard deviations are reported. These protocols are kept the same for all methods to guarantee fair comparisons. We summarize the pipeline of unsupervised settings for \model in Algorithm~\ref{algo:pipeline_un}.

\begin{algorithm}
\caption{The pipeline of unsupervised settings for \model} 
\label{algo:pipeline_un}
\begin{algorithmic}[1]
\REQUIRE Graph $\mathcal{G}$ without labels, training epochs $L$.
\STATE Construct the dientangled graph architecture super-networks with randomly initialized weights $w$ and operation choices $\alpha$.
\FOR{$l = 1, \dots, L$}
    \STATE Calculate the self-supervised training loss with architecture-graph disentanglement $\mathcal{L}_w$ as Eq. (9)
    \STATE Update the super-network weights with $w = w-\lambda_w\nabla_w \mathcal{L}_w$
    \STATE Calculate the contrastive search loss with architecture augmentations $\mathcal{L}_\alpha$ as Eq. (12)
    \STATE Update the super-network operation choices with $\alpha = \alpha - \lambda_\alpha \nabla_\alpha \mathcal{L}_{\alpha}$
\ENDFOR
\STATE Evaluate the searched model with linear protocols.
\end{algorithmic}
\end{algorithm}

\subsection{Semi-supervised Settings}
\paragraph{Setups} To further test the performance of GNAS in scenarios with scarce labels instead of exactly no labels, we conduct semi-supervised experiments with limited labels, i.e., using 10\%,5\%,1\% labels for both training and validation. In this setting, we compare the differentiable GNAS baselines, including DARTS, PAS and GASSO, which train super-network weights with training datasets and optimize architecture parameters with validation datasets as in supervised settings. For \modelnosp, it first pretrains the super-network by methods mentioned in Section 3.2 and Section 3.3, and then continues the search process as traditional supervised GNAS. We also include DSGAS-P, which does not adopt the pretraining stage, as an ablated baseline. 

\paragraph{Evaluation protocols} For OGBG-Molhiv and OGBN-Arxiv, the splits are the same in the open graph benchmark~\cite{hu2020open}. For Wechat-Video, we adopt random splits with a ratio of 6:2:2 for training, validation, and testing by multi-label stratified splitting~\cite{sechidis2011stratification}. The available training and validation labels are randomly sampled with a stratified sampling for settings of labeling rates 1\%, 5\%, and 10\%. We train the models with early-stop patience 50, and then we adopt the best-performed checkpoint on validation data split, which is tested on testing data split to obtain the reported results. These splits and the training strategies are kept the same for all methods to guarantee fair comparisons. The experiments are run five times with different random seeds. We summarize the pipeline of semi-supervised settings for \model in Algorithm~\ref{algo:pipeline_semi}.

\begin{algorithm}
\caption{The pipeline of semi-supervised settings for \model} 
\label{algo:pipeline_semi}
\begin{algorithmic}[1]
\REQUIRE Graph $\mathcal{G}$ with limited labels, training epochs $L$, earlystop patience $E$.
\STATE Construct the dientangled graph architecture super-networks with randomly initialized weights $w$ and operation choices $\alpha$.
\STATE Pretraining the super-networks for $w$ and $\alpha$ using Algorithm~\ref{algo:pipeline_un} 
\FOR{$l = 1, \dots, L$}
    \STATE Calculate the supervised loss on training data $\mathcal{L}_w$.
    \STATE Update the super-network weights with $w = w-\lambda_w\nabla_w \mathcal{L}_w$.
    \STATE Calculate the supervised loss on validation data $\mathcal{L}_\alpha$.
    \STATE Update the super-network operation choices with $\alpha = \alpha - \lambda_\alpha \nabla_\alpha \mathcal{L}_{\alpha}$.
    \IF{ the validation accuracy is non-increasing for $E$ epochs}
    \STATE break
    \ENDIF
\ENDFOR
\STATE Evaluate the searched model on testing data.
\end{algorithmic}
\end{algorithm}

\section{Implementation Details}
\subsection{Super-network Construction}
The super-network generally consists of two parts, the operation pool and the directed acyclic graph (DAG) that wires the operations. Following ~\cite{wei2021pooling}, we adopt three kinds of operations in the operation pool as follows:
\begin{itemize}[leftmargin = 0.5cm]
    \item Node aggregation operations, which aggregate messages from the neighborhood to update the node representations, including GCN~\cite{kipf2016semi}, GAT~\cite{velivckovic2018graph}, GIN~\cite{xu2018powerful}, GraphConv~\cite{morris2019weisfeiler}, GraphSage~\cite{hamilton2017inductive}. MLP (Multi-layer Perceptrons) is also included as an operation that does not utilize the neighborhood. 
    \item Graph pooling operations, which aggregate node representations to obtain graph-level representations, including SortPool~\cite{zhang2018end}, AttentionPool~\cite{li2015gated}, MaxPool, MeanPool and SumPool. For example, MeanPool takes the average of the node representations as the graph representation. 
    \item Layer merging operations, which aggregate representations from intermediate layers to formulate more expressive representations, including MaxMerge, ConcatMerge, SumMerge and MeanMerge. For example, MaxMerge selects the max values in multiple representations from intermediate layers. 
\end{itemize}

For brevity, we denote `Agg',`Pool',`Merge' as node aggregation operations, graph pooling operations, and layer merging operations respectively. Following ~\cite{qin2021graph}, the DAG for node classification tasks is a straightforward path, i.e., $\mathbf{H}^{l+1} = \text{Agg}^{l}(\mathbf{H}^{l},\mathbf{A})$, and the embeddings of the last layer are utilized for downstream tasks, where $\mathbf{H}^{l}$ denotes the hidden embeddings output by the $l$-th layer, and $\mathbf{A}$ denotes the graph adjacency matrix. Following ~\cite{wei2021pooling}, the DAG for graph classification tasks is constructed by $\mathbf{H}^{l+1} = \text{Agg}^{l}(\mathbf{H}^{l},\mathbf{A}), \mathbf{Z}^{l} = \text{Pool}^{l}(\mathbf{H}^{l},\mathbf{A})$, and the merged representations $\text{Merge}(\mathbf{Z}^{1},\mathbf{Z}^{2},\dots,\mathbf{Z}^{L})$ are utilized for downstream tasks, where $L$ is the number of layers. 

\subsection{Hyperparameters}
For fair comparisons, all methods adopt the same dimensionality, number of layers and normalization techniques.  For graph classification datasets, we adopt the dimensionality as 32, the number of layers as 3 and batch normalization~\cite{ioffe2015batch}. For node classification datasets, we adopt the dimensionality as 128 and the number of layers as 2 and layer normalization~\cite{ba2016layer}. Adam optimizer~\cite{kingma2014adam} is adopted to optimize the model weights and another SGD optimizer is adopted to optimize architecture parameters for NAS methods. For our method, we adopt $K=3$ for all node classification datasets and $K=4$ for all graph classification datasets, and the hyperparameters that control the perturbation degree for the architecture augmentations of operation choices, weights and embeddings are set to 1.1, 0.1, 0.05 respectively for all datasets.

\subsection{Configurations}
All experiments are conducted with:
\begin{itemize}[leftmargin=0.5cm]
    \item Operating System: Ubuntu 20.04.5 LTS
    \item CPU: Intel(R) Xeon(R) Gold 6240 CPU @ 2.60GHz
    \item GPU: NVIDIA GeForce RTX 3090 with 24 GB of memory
    \item Software: Python 3.9.12, Cuda 11.3, PyTorch~\cite{paszke2019pytorch} 1.12.1, PyTorch Geometric~\cite{Fey/Lenssen/2019} 2.0.4.
\end{itemize} 

\end{document}